\documentclass[10pt,twocolumn,letterpaper]{article}

\usepackage[pagenumbers]{cvpr} 

\definecolor{cvprblue}{rgb}{0.21,0.49,0.74}
\usepackage[pagebackref,breaklinks,colorlinks,allcolors=cvprblue]{hyperref}

\def\confName{CVPR}
\def\confYear{2027}

\usepackage[utf8]{inputenc} 
\usepackage[T1]{fontenc}    
\usepackage{url}            
\usepackage{booktabs}       
\usepackage{amsfonts}       
\usepackage{nicefrac}       
\usepackage{microtype}      
\usepackage{xcolor}         

\usepackage{wrapfig}
\usepackage{multirow}
\usepackage{amsmath, amsthm}
\usepackage{amssymb}
\usepackage[pagebackref,breaklinks,colorlinks,allcolors=cvprblue]{hyperref}
\usepackage[nameinlink, capitalize]{cleveref}
\usepackage{algorithm}
\usepackage{algpseudocode}
\usepackage{graphicx}
\usepackage{float}
\usepackage{siunitx}
\usepackage[table]{xcolor}  
\usepackage{array}
\usepackage{stackengine}
\usepackage{pifont}
\definecolor{diffblue}{HTML}{0000FF}

\theoremstyle{plain}
\newtheorem{theorem}{Theorem}[section]

\theoremstyle{definition}
\newtheorem{definition}[theorem]{Definition}

\theoremstyle{remark}

\newtheorem{innercustomthm}{Theorem}
\newenvironment{customthm}[1]
  {\renewcommand\theinnercustomthm{#1}\innercustomthm}
  {\endinnercustomthm}

\newtheorem{innercustomdef}{Definition}
\newenvironment{customdef}[1]
  {\renewcommand\theinnercustomdef{#1}\innercustomdef}
  {\endinnercustomdef}

\usepackage[many]{tcolorbox}
\tcbuselibrary{theorems}

\title{Temporal Predictive Coding for Training-Free ANN-to-SNN Conversion}

\author{
Xiaoli Liu \textsuperscript{1} \quad\quad
Malu Zhang \textsuperscript{1} \thanks{Corresponding author} \quad\quad
Yang Yang \textsuperscript{1} \\
\textsuperscript{1}School of Computer Science and Engineering, \\ University of Electronic Science and Technology of China\\
{\ttfamily\small
liuxiaoli0720@std.uestc.edu.cn
\quad maluzhang@uestc.edu.cn}
}

\begin{document}
\maketitle
\begin{abstract}
Spiking neural networks (SNNs) hold considerable promise owing to their low power consumption. Converting artificial neural networks (ANNs) into spiking counterparts provides an effective route to high-performance SNNs. However, the rate-coding schemes employed by many existing conversion methods necessitate more time steps than directly trained SNNs, resulting in increased energy consumption. In this work, we propose a brain-inspired Temporal Predictive Coding (TPC) framework for ANN-to-SNN conversion. TPC leverages temporal dynamics to predict the current state of neurons and treats discrete spikes purely as corrections to prediction errors. This approach improves the network's conversion accuracy and significantly reduces the required time-steps. Furthermore, we theoretically analyze the expected conversion error and derive the optimal analytical decision boundary for multi-threshold quantization. Based on this analysis, we incorporate a damping control mechanism to prevent error accumulation in deep networks. Experimental results across vision (CNNs and Transformers), language (BERT) and multimodal (Vision-Language Models) tasks demonstrate that the proposed TPC framework significantly improves accuracy while reducing time-steps and energy consumption, achieving state-of-the-art performance.
\end{abstract}    
\section{Introduction}
\label{sec: introduction}


Spiking neural networks (SNNs) \cite{maass1997networks,izhikevich2003simple,masquelier2008spike} are commonly recognized as third-generation neural network models, distinguished by their unique neuronal dynamics and high biological plausibility. These characteristics position them as a competitive alternative to artificial neural networks (ANNs) \cite{li2024brain}. A key difference between the two lies in the representation of information \cite{zhou2022spikformer,yao2023spike}: ANNs pass dense floating-point values from one layer to the next, whereas SNNs communicate through sparse binary spikes generated when neuronal membrane potentials cross a firing threshold \cite{zhou2022spikformer,xing2024spikelm,xing2024spikellm}. This spike-based signaling supports event-driven computation \cite{zhu2022training,zhang2023self} and enables extremely low power consumption when implemented on neuromorphic chips \cite{davies2018loihi,akopyan2015truenorth,pei2019towards}.

\begin{figure}[htbp]
    \centering
    \includegraphics[width=\linewidth]{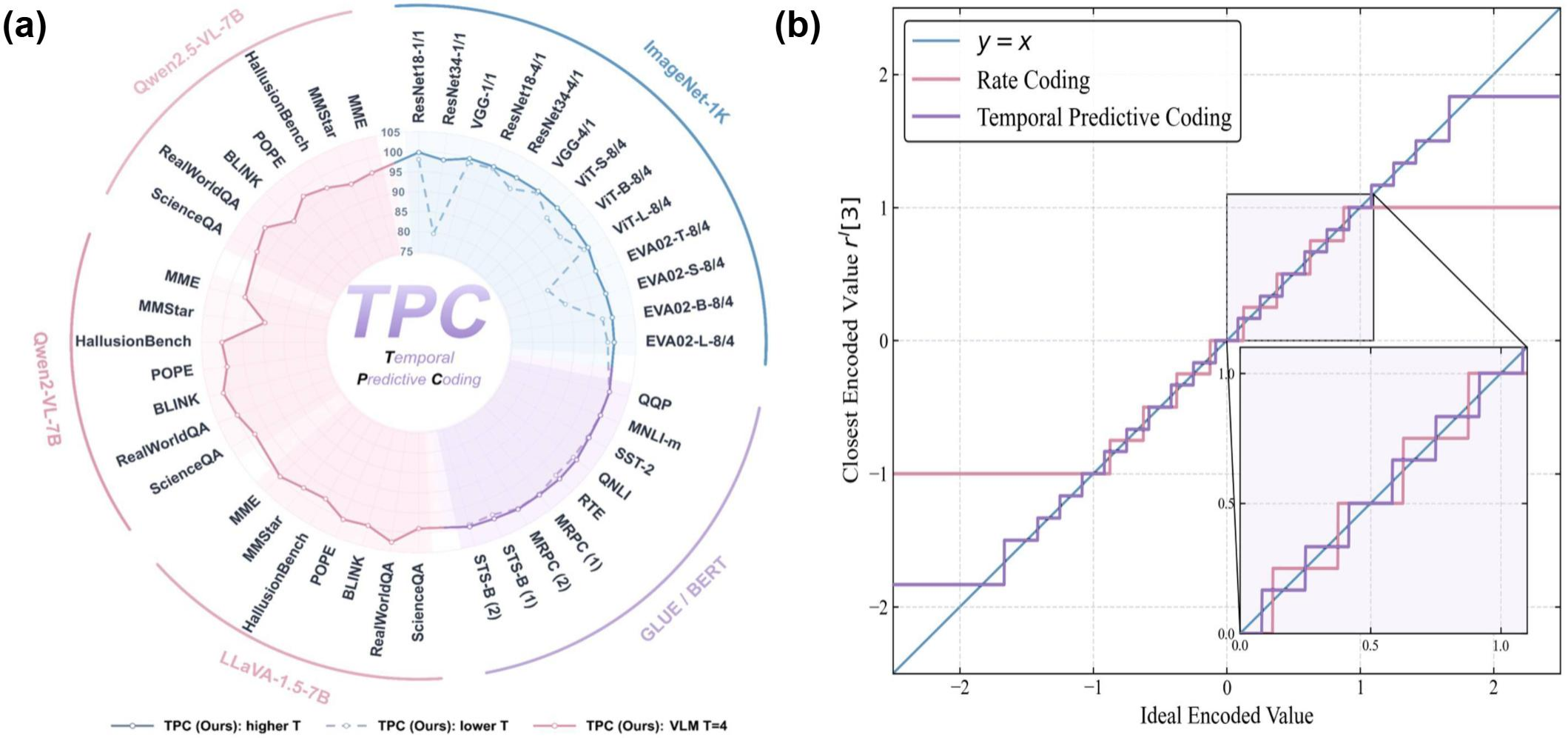}
    \vspace{-4mm}
    \caption{(a) Normalized TPC performance across vision, language, and vision–language benchmarks. (b) Comparison of ideal fitting results for input y = x. TPC shows a wider representation range and higher precision.}
    \vspace{-6mm}
    \label{fig:teaser}
\end{figure}



The discrete and non-differentiable nature of spike generation, however, makes SNN training particularly challenging. Surrogate gradient methods \cite{wu2018spatio} partially overcome this obstacle by using the derivatives of differentiable surrogate functions in place of spike-function derivatives during backpropagation. This strategy enables deep SNNs to be trained through backpropagation and gradient descent, substantially improving their performance \cite{wu2019direct,zheng2021going,yao2024spike,yao2025scaling}. Despite these advances, surrogate gradients remain coarse approximations and may produce inaccurate descent directions in multilayer SNNs. Some online learning methods \cite{xiao2022online,xiao2024online} estimate the full backpropagation-through-time (BPTT) gradient by aggregating single-step gradients, but their task accuracy remains suboptimal.

\begin{figure*}[htbp]
    \centering    
    \includegraphics[width=1.0\textwidth]{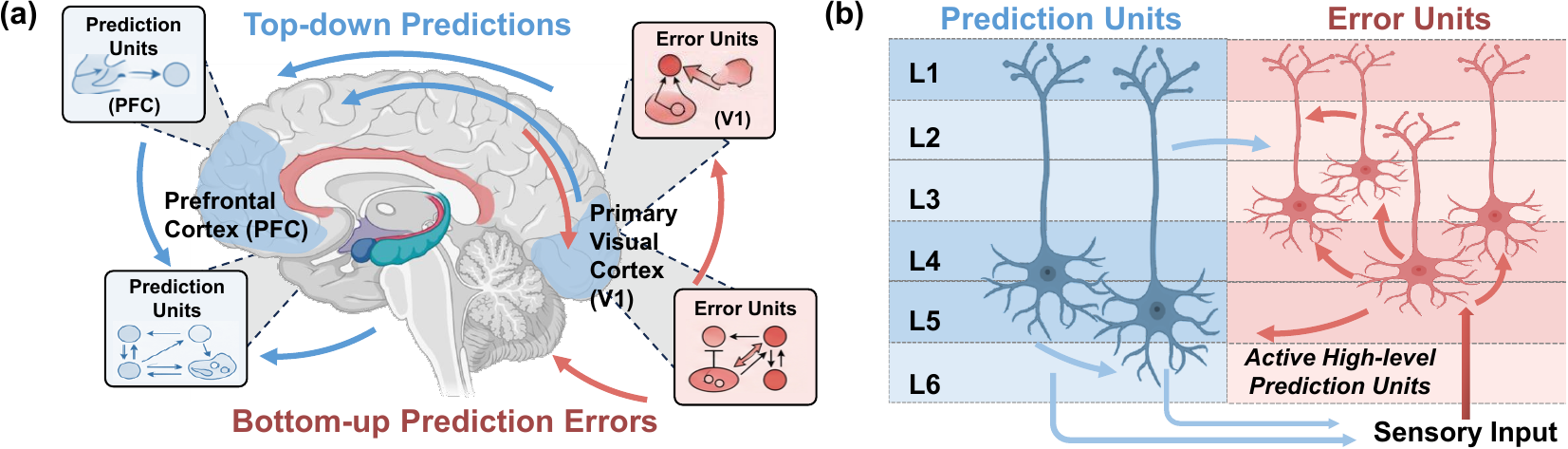}  
    \vspace{-5mm}
    \caption{(a) Hierarchical processing and information flow; (b) Hierarchical cortical circuits.}
    \vspace{-5mm}
    \label{fig:neuron and predictive coding}
\end{figure*}

In addition to surrogate gradient, ANN-to-SNN conversion methods \cite{cao2015spiking,rueckauer2017conversion,deng2021optimal,li2021free} represent another class of spiking deep learning methodologies designed to address the challenges of training SNNs. These methods replace nonlinear activation functions with spiking neurons to convert pre-trained ANNs into SNNs \cite{bu2022optimized}. Most conversion methods are based on rate coding \cite{kim2022rate}, which represents the activations in ANNs through the firing rates of SNNs. However, accurately estimating firing rates requires a large number of time-steps, resulting in conversion methods having higher energy consumption than surrogate gradient methods.

Rethinking the brain’s mechanisms, we find that traditional rate coding violates the principles of efficient information processing in neural systems. In neuroscience, the theory of predictive coding \cite{friston2018does} suggests that in order to minimize energy consumption \cite{ali2022predictive,zhang2025energy}, the brain does not continuously transmit absolute sensory signals between neural layers. Instead, higher-order cortical regions send an internal prediction to lower-level layers, and neurons in those lower layers transmit upward only the residual difference between the actual input and the prediction—that is, the prediction error—in the form of sparse spikes \cite{rao1999predictive}. This biological mechanism, based on temporal inertia and adaptability, enables the brain to filter out redundant information and instantly capture dynamic changes in the environment using a minimal number of spikes.

Inspired by this mechanism, we propose a Temporal Predictive Coding (TPC) framework for conversion. Unlike traditional approaches that treat the average firing rate as the encoded activation value, TPC utilizes temporal dynamics to predict the current state of neurons and treats discrete spikes as precise corrections to prediction errors. This method not only improves the network's conversion accuracy but also reduces the number of time-steps required, thereby lower energy consumption without the need for additional training. Furthermore, we theoretically analyze the expected conversion error and derive the optimal analytical decision boundary for multi-threshold quantization. Based on this analysis, to prevent error accumulation in deep networks, we incorporate a damping control mechanism to ensure rapid convergence under low time-steps. Our main contributions are summarized as follows:
\begin{itemize}
    \item We propose Temporal Predictive Coding (TPC), a novel framework for ANN-to-SNN conversion. TPC predicts neuron states via temporal dynamics and transmits prediction errors using sparse spikes. This overcomes the redundancy of traditional rate coding, which relies on transmitting absolute values.
    \item We theoretically analyze the expected conversion error and derive the analytically optimal decision boundary for the multi-threshold quantization function. Based on this, we introduce a damping control mechanism to suppress residual error accumulation in deep networks, ensuring rapid convergence under short time-steps.
    \item Experimental results across vision (CNNs and Transformers), language (BERT) and multimodal (Vision-Language Models) tasks demonstrate that the proposed TPC framework significantly improves accuracy while reducing time-steps and energy consumption, achieving state-of-the-art performance.
\end{itemize}
\section{Related Works}
\subsection{Rate Coding in ANN-to-SNN Conversion}

\begin{figure*}[htbp]
    \centering    
    \includegraphics[width=1.0\textwidth]{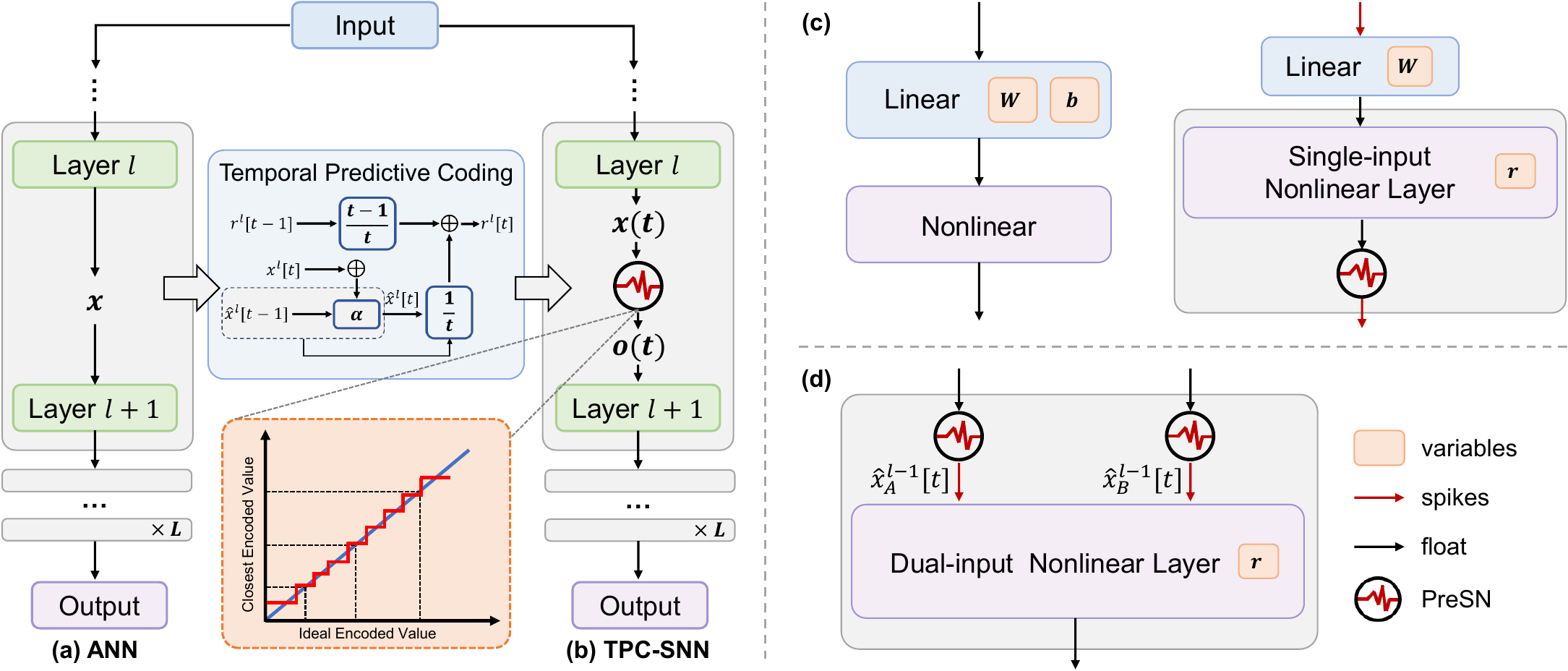}  
    \vspace{-5mm}
    \caption{The framework of our method. (a) The standard ANN; (b) The converted TPC-SNN. (c) Conversion of a linear layer followed by a nonlinear layer in the ANN into SNN modules. (d) Conversion of dual-input nonlinear layer in the ANN into SNN modules.}
    \vspace{-5mm}
    \label{fig:main figure}
\end{figure*}

Rate coding is one of the most commonly used conversion methods. It was first observed in biological neural systems \cite{adrian1926impulses}, where stronger stimuli lead to more frequent firing. This method bridges the gap between the ReLU activation in ANNs and the firing rate of IF neurons in SNNs. Based on this, the main conversion methods have been derived \cite{cao2015spiking}. Furthermore, the normalization of weights for regulating activations \cite{rueckauer2017conversion} and the balancing of thresholds for spiking neurons \cite{han2020rmp} have been proposed to reduce the mismatch range during the conversion. However, the floating-point activations of the source ANNs are continuous, and discrete firing rates do not align well with them, leading to quantization errors. To reduce conversion errors, several methods for quantizing ANNs to SNNs have been proposed \cite{buoptimal,hu2023fast}. During inference, spikes may not arrive uniformly, which can lead to non-uniform errors in the conversion \cite{buoptimal}. Several methods have been proposed to reduce this error, including a two-stage inference strategy \cite{hao2023reducing} and shifting the initial membrane potential \cite{haobridging}. Recent research has expanded to converting ANNs beyond ReLU into SNNs. \cite{oh2024sign} introduced a method based on sign gradient descent applied to neurons, which can approximate various nonlinear activation functions. \cite{wang2023masked} and \cite{you2024spikezip} trained modified Transformers and converted them into SNNs. \cite{jiang2024spatio} and \cite{huang2024towards} developed modules approximating nonlinear layers, enabling training-free conversion from Transformers to SNNs.

\subsection{Predictive Coding in Neuroscience}

In computational neuroscience, traditional perspective often model sensory information processing as a passive, feedforward procedure. In deep learning, this view has evolved into the transmission of dense absolute values across layers in ANNs, and the representation of these absolute values via firing rates in SNNs (i.e., rate coding) \cite{diehl2015fast}. However, predictive coding \cite{rao1999predictive,friston2018does} introduces a fundamentally different and highly efficient mechanism, proposing that the brain operates essentially as a proactive inference system \cite{friston2010free}.

The core principle of predictive coding is the minimization of communication redundancy and energy consumption within the system \cite{ali2022predictive,barlow1961possible}. Under this mechanism, the neural system refrains from continuously transmitting redundant absolute sensory signals across hierarchical layers. Instead, higher-order neurons generate and send predictions of sensory inputs to lower layers, while lower-order neurons compute only the residual difference between the actual input and the prediction \cite{bastos2012canonical}, known as the prediction error (\cref{fig:neuron and predictive coding}(b)). Within this framework, discrete spikes no longer function as counters accumulating absolute values; rather, they serve purely as asynchronous error-correction signals \cite{deneve2016efficient}. When the variation in input signals aligns with the neuron's internal prediction, the neuron remains in a resting state, thereby filtering out redundant information. A neuron fires sparse spikes to precisely relay the prediction error to the subsequent layer only when a deviation between the prediction and the actual state occurs. This mechanism not only endows the brain with activation sparsity but also provides an ideal theoretical paradigm for ANN-to-SNN conversion.

\section{Methodology}
\label{section: methodology}

\subsection{Temporal Predictive Coding Framework}
\label{methods: tpc framework}

The ANN-to-SNN conversion often uses rate coding to transmit information, where the firing rate \(r^l[t]\) at each time-step encodes the activation value. \cref{equ: rate coding} shows the relationship between the output firing rate \(r^l[t]\) and the output signal \(x^l[t]\). When layer \(l\) consists of spiking neurons, the state can be described as \(x^l[t]=\theta^l s^l[t]\), where \(s^l[t]\) denotes the spike at time-steps \(t\) and \(\theta^l\) represents the threshold.
\begin{equation}
\label{equ: rate coding}
    r^l[t]=\frac{1}{t}\sum^T_{t=1}x^l[t]=\frac{t-1}{t}r^l[t-1]+\frac{x^l[t]}{t}.
\end{equation}

However, the rate coding has a problem: accurately estimating the firing rate requires a large number of time-steps, resulting in significantly higher latency and energy consumption for the conversion method compared to the surrogate gradient method. To address this issue, we propose a novel brain-inspired encoding scheme, referred to as \textbf{\textit{Temporal Predictive Coding}} (\cref{fig:main figure}).

\begin{definition}
\label{def: temporal predictive coding}
In TPC, denote \(x^l[t]\) as the actual transmitted prediction error of the neuron at time-step \(t\). Define \(\hat{x}^l[t]\) as the reconstructed raw differential output, and \(r^l[t]\) as the encoded continuous activation value at time-step \(t\). The relationship among them is expressed as follows:
\begin{align}
\label{equ: tpc framework}
    & \hat{x}^l[t]=x^l[t]+\mathcal{P}_t, \\
    & r^l[t]=\frac{t-1}{t}r^l[t-1]+\frac{\hat{x}^l[t]}{t},
\end{align}
where \(\mathcal{P}_t\) represents the damped temporal predictive function based on historical momentum, \(t\) starts from \(1\), and \(r^l[0]=0\) (\textit{proof in \cref{supp: proof of the definition 4.1.}}).
\end{definition}

In addition, another issue with rate coding is that the encoded values gradually decay over time. Unlike rate coding which suffers from progressive decay, TPC leverages \(\mathcal{P}(\cdot)\) to predict the current state based on temporal inertia. The actual transmitted signal \(x^l[t]\) is purely the asynchronous error correction. The encoded activation \(r^l[t]\) matches the target continuous activation via temporal expectation tracking, seamlessly integrating the predictive coding mechanism into ANN-to-SNN conversion. As shown in \cref{fig:teaser}(b), TPC can represent a wider range of values and achieve higher precision than rate coding at the same threshold and time-steps.

\begin{figure}[htbp]
    \centering
    \includegraphics[width=\linewidth]{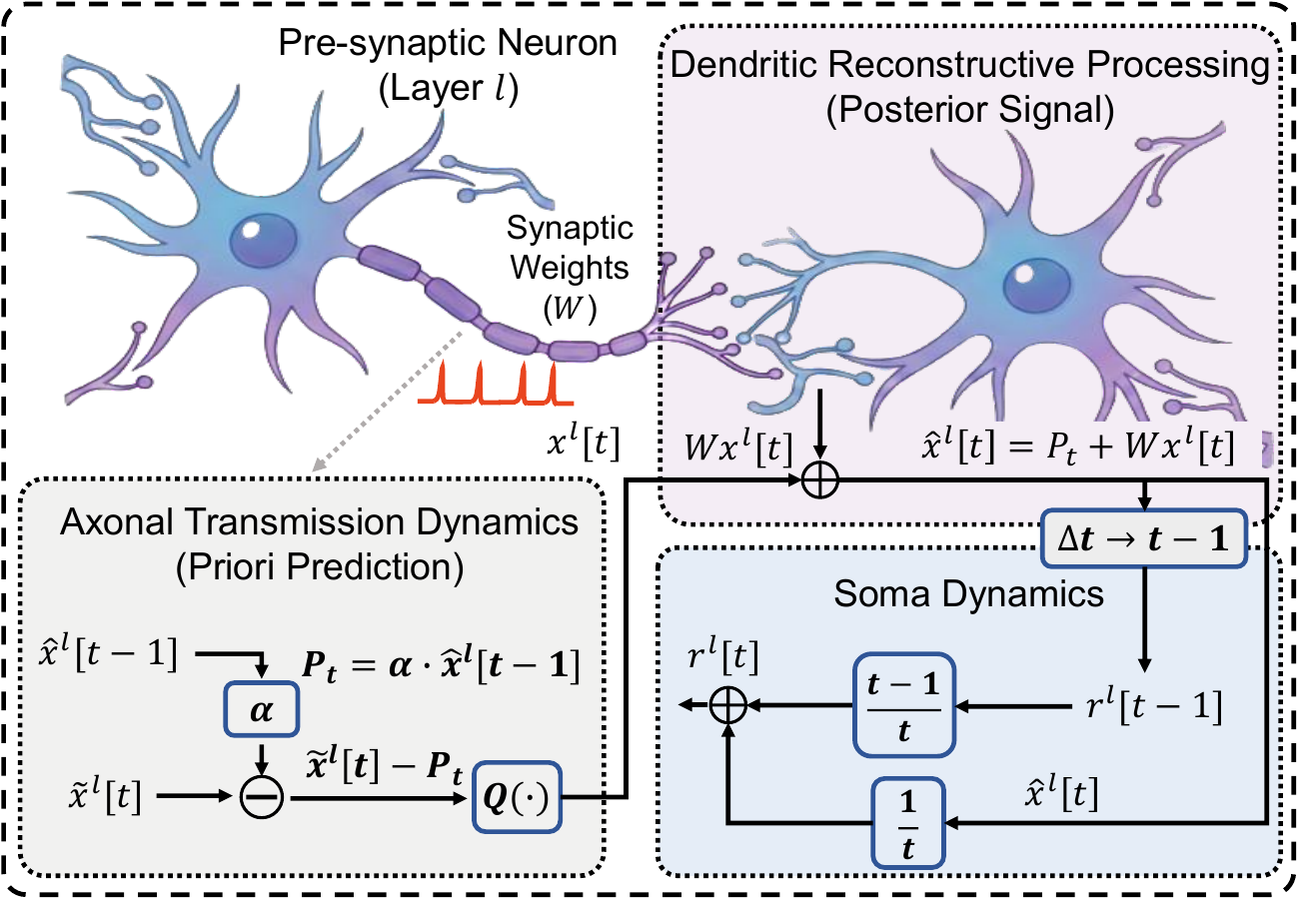}
    \vspace{-5mm}
    \caption{Dynamics of the Predictive Spiking Neurons.}
    \vspace{-5mm}
    \label{fig:tpc neuron}
\end{figure}

\subsection{Predictive Spiking Neurons}
\label{methods: predictive neurons with MTQ}

Traditional spiking neurons (such as the IF neurons in rate coding) approximate continuous activation values by passively accumulating input currents. Within the TPC framework, the dynamics of neurons are redefined as active expectation error trackers (\cref{fig:tpc neuron}). At time-step \(t\), predictive spiking neurons (PreSNs) do not directly transmit complete absolute activation values, but instead follow a “predict-update” temporal dynamic. This predict-update dynamic mathematically parallels the recursive state estimation in Kalman Filter \cite{kalman1960new,welch1995introduction} (see detailed discussion in \cref{supp: discussion}). Based on multi-threshold spiking neuron \cite{huang2024towards} (refer to \cref{supp: multi-threshold spiking neuron}), PreSNs utilize an internal damping mechanism to extract temporal inertia and, based on the posterior reconstruction signal \(\hat{x}^l[t-1]\) from the previous time-step, generate a prior prediction \(\mathcal{P}_t\) of the current activation state:
\begin{equation}
\label{equ: dampling control}
    \mathcal{P}_t=\alpha \cdot \hat{x}^l[t-1],
\end{equation}
where \(\alpha \in [0,1)\) is the damping coefficient. Based on this prior prediction, at the transmitter, PreSNs computes the ideal target signal for this layer, \(\tilde x^l[t]\). After stripping away the prior prediction, it compresses the signal using a multi-threshold quantization function \(\mathcal{Q}(\cdot)\) into a discrete spike residual \(x^l[t]\), which is then transmitted to the next layer:
\begin{equation}
\label{equ:quantization on spike residual}
    x^l[t] = \mathcal{Q}\big( \tilde{x}^l[t] - \mathcal{P}_t \big).
\end{equation}

At the receiver, since the sender and receiver share the same temporal dynamics, the network superimposes the received \(x^l[t]\) onto the internal predicted state \(\mathcal{P}_t\) to reconstruct the current posterior signal \(\hat x^l[t]\): \(\hat{x}^l[t] = \mathcal{P}_t + x^l[t]\). Finally, by performing a strict time-weighted sliding integration on the posterior reconstructed signal, the neuron continuously updates and tracks the steady-state expectation \(r^l[t]\) of the current feature:
\begin{equation}
    r^l[t] = \frac{t-1}{t}r^l[t-1] + \frac{\hat{x}^l[t]}{t}.
\end{equation}

\cref{thm: neurons in rate coding to presns} demonstrates how to convert spiking neurons in rate coding to PreSNs in temporal predictive coding. 

\begin{theorem}
\label{thm: neurons in rate coding to presns}
In rate coding, the output spike \(s^{l-1}[t]\) from the previous layer generates the input current to the current layer via the synaptic weight \(W^l\), i.e., \(I^l[t] = W^l s^{l-1}[t]\). Within the TPC framework, by introducing a local state \(m^l[t]\), any spiking neuron can be equivalently converted into a PreSN. The dynamical update equation for its equivalent membrane current \(I^l[t]\) is as follows:
\begin{align}
    & I^l[t]=\begin{cases}
        W^lx^{l-1}[t]+b^l, \ t=1 \\
        W^lx^{l-1}[t]+\alpha \cdot m^l[t], \ t>1
    \end{cases} \\
    & m^{l}[t+1]=\alpha \cdot m^l[t]+W^l x^{l-1}[t]-x^l[t],
\end{align}
where \(x^{l-1}[t]\) is the output of the previous layer, \(b^l\) is injected only at the first time-step, and the local state is initialized as \(m^l[1]=0\) (Proof in \cref{supp: proof of the theorem 4.2.}).
\end{theorem}

\subsection{TPC Conversion for Linear Layers}

\cref{thm: tpc for linear layers} illustrates the equivalent transformations of linear layers within the TPC framework.

\begin{theorem}
\label{thm: tpc for linear layers}
For linear layers, including fully connected layers and convolutional layers, which can be expressed as \cref{equ: linear layer},
\begin{equation}
\label{equ: linear layer}
    r^l_{\text{ANN}} = W^l_{\text{ANN}} r^{l-1}_{\text{ANN}} + b^l_{\text{ANN}},
\end{equation}
where \(W^l\) and \(b^l\) represent the weights and biases of the \(l\)-th layer, respectively. Within the TPC framework, the conversion for a linear layer is equivalent to eliminating the static bias term \(b^l\) from the forward computation and incorporating the bias term into the initial state of the subsequent PreSN at the first time-step:
\begin{equation}
\hat{x}^l[t] = 
\begin{cases} W^l \hat{x}^{l-1}[t] + b^l, & \text{if } t = 1 \\ 
W^l \hat{x}^{l-1}[t], & \text{if } t > 1 
\end{cases}
\end{equation}
\end{theorem}
The proof of \cref{thm: tpc for linear layers} is detailed in the \cref{supp: proof of the theorem 4.3.}.

\subsection{TPC Conversion for Nonlinear Layers}

Based on Definition \cref{def: temporal predictive coding},We introduce the principle of expectation domain equivalence and apply it to the conversion of nonlinear layers. \cref{thm: tpc for single-input nonlinear layers} (Proof in \cref{supp: proof of the theorem 3.4.}) corresponds to a nonlinear layer with a single input \(x^{l-1}[t]\), while \cref{thm: tpc for multi-input nonlinear layers} (Proof in \cref{supp: proof of the theorem 3.5.}) applies to a nonlinear layer with two inputs, \(x^{l-1}_A[t]\) and \(x^{l-1}_B[t]\).

\begin{theorem}
\label{thm: tpc for single-input nonlinear layers}
Let \(f\) be a nonlinear layer of order \(l\) that takes only a single input \(x^{l-1}[t]\), such as ReLU, SiLU, MaxPool, LayerNorm, or Softmax. In the TPC framework, if the reconstructed signal \(\hat x^l[t]\) generated by this layer at time \(t\) satisfies the following equation:
\begin{equation}
    \hat{x}^l[t] = t \cdot f(r^{l-1}[t]) - (t-1) \cdot f(r^{l-1}[t-1]),
\end{equation}
then, after any \(t\) time-steps, the steady-state expectation \(r^l[t]\) of the output from this nonlinear layer equals the absolute nonlinear activation value of the source ANN, i.e., \(r^l[t] = f(r^{l-1}[t])\).
\end{theorem}

\begin{theorem}
\label{thm: tpc for multi-input nonlinear layers}
Let \(\otimes\) be an operator with two inputs, \(x^{l-1}_A[t]\) and \(x^{l-1}_B[t]\), such as matrix multiplication or element-wise multiplication. In the ANN-to-SNN conversion, the mapping from the operator \(\otimes\) to the TPC dynamics is given by:
\begin{align}
\label{equ: tpc for multi-input nonlinear layer1}
& r_A^{l-1}[t] = \frac{t-1}{t}r_A^{l-1}[t-1] + \frac{\hat{x}_A^{l-1}[t]}{t},  \\
\label{equ: tpc for multi-input nonlinear layer2}
& r_B^{l-1}[t] = \frac{t-1}{t}r_B^{l-1}[t-1] + \frac{\hat{x}_B^{l-1}[t]}{t}, \\ 
\label{equ: tpc for multi-input nonlinear layer3}
\begin{split}
&\hat{x}^l[t] = \frac{1}{t} \big( \hat{x}_A^{l-1}[t] \otimes \hat{x}_B^{l-1}[t] \big) + \frac{t-1}{t} \Big( \hat{x}_A^{l-1}[t] \otimes r_B^{l-1}[t-1] \\
    &\quad + r_A^{l-1}[t-1] \otimes \hat{x}_B^{l-1}[t] - r_A^{l-1}[t-1] \otimes r_B^{l-1}[t-1] \Big),
\end{split}
\end{align}
where \(\hat x^{l-1}_A[t]\) and \(\hat x^{l-1}_B[t]\) are the posterior reconstruction signals of the previous layer received at time step \(t\), respectively; \(r_A^{l-1}[t]\) and \(r_B^{l-1}[t]\) are the steady-state expectations of the internal tracking. The output of layer \(l\) at time-step \(t\) (which serves as the input to layer \(l+1\)) is given by \(x^l[t] = \mathcal{Q}\big(\hat{x}^l[t] - \alpha \cdot \hat{x}^l[t-1]\big)\).
\end{theorem}

\begin{table*}[htbp]
  \centering
  \caption{Comparison between the proposed method and previous works on ImageNet-1K.}
  \vspace{-2mm}
  \label{tab:imagenet}
  \resizebox{\textwidth}{!}{
  \begin{tabular}{ccccccc}
    \toprule
    \textbf{Method} & \textbf{Type} & \textbf{Arch} & \textbf{Param (M)} & \textbf{ANN Acc (\%)} & \(T\) & \textbf{SNN Acc (\%)} \\
    \midrule  
    \multirow{1}{*}{TS \cite{deng2021optimal}} & \multirow{1}{*}{CNN-to-SNN} & VGG-16 & 138 & 72.40 & 64 & 70.97 \\
    \midrule
    \multirow{1}{*}{SNM \cite{wang2022signed}} & \multirow{1}{*}{CNN-to-SNN} & VGG-16 & 138 & 73.18 & 64 & 71.50 \\
    \midrule
    \multirow{2}{*}{MMSE \cite{li2021free}} & \multirow{2}{*}{CNN-to-SNN} & ResNet-34 & 21.8 & 75.66 & 64 & 71.12 \\
     & & VGG-16 & 138 & 75.36 & 64 & 70.69 \\
    \midrule
    \multirow{2}{*}{QCFS \cite{buoptimal}} & \multirow{2}{*}{CNN-to-SNN} & ResNet-34 & 21.8 & 74.32 & 64 & 72.35 \\
     & & VGG-16 & 138 & 74.29 & 64 & 72.85 \\
    \midrule
    \multirow{2}{*}{SRP \cite{hao2023reducing}} & \multirow{2}{*}{CNN-to-SNN} & ResNet-34 & 21.8 & 74.32 & 4, 64 & 66.71, 68.61 \\
     & & VGG-16 & 138 & 74.29 & 4, 64 & 66.47, 69.43 \\
    \midrule
    \multirow{1}{*}{MST \cite{wang2023masked}} & Transformer-to-SNN & Swin-T (BN) & 28.5 & 80.51 & 128, 512 & 77.88, 78.51 \\
    \midrule
    \multirow{1}{*}{STA \cite{jiang2024spatio}} & Transformer-to-SNN & ViT-B/32 & 86 & 83.6 & 32, 256 & 78.72, 82.79 \\
    \midrule
    \multirow{3}{*}{SpikeZIP-TF \cite{you2024spikezip}} & \multirow{3}{*}{Transformer-to-SNN} & SViT-S-32Level & 22.05 & 81.59 & 64 & 81.45 \\
     & & SViT-B-32Level & 86.57 & 82.83 & 64 & 82.71 \\
     & & SViT-L-32Level & 304.33 & 83.86 & 64 & 83.82 \\
    \midrule
    \multirow{2}{*}{ECMT \cite{huang2024towards}} & \multirow{2}{*}{Transformer-to-SNN} & ViT-S/16 & 22 & 78.04 & 8, 10 & 76.03, 77.07 \\
     & & EVA-G & 1074 & 89.62 & 4, 8 & 88.60, 89.40 \\
    \midrule
    \multirow{13}{*}{DCGS \cite{huang2025differential}} & \multirow{6}{*}{CNN-to-SNN} & ResNet18-1/1 & 11.7 & 71.49 & 32, 64 & 69.89, 71.08 \\
     & & ResNet34-1/1 & 21.8 & 76.42 & 32, 64 & 58.86, 74.11 \\
     & & VGG-1/1 & 138 & 73.25 & 32, 64 & 72.04, 73.13 \\
     & & ResNet18-4/1 & 11.7 & 71.49 & 4, 8 & 70.07, 71.31 \\
     & & ResNet34-4/1 & 21.8 & 76.42 & 4, 8 & 73.35, 76.04 \\
     & & VGG-4/1 & 138 & 73.25 & 4, 8 & 72.72, 73.17 \\
     \cmidrule{2-7}
     & \multirow{7}{*}{Transformer-to-SNN} & ViT-S-8/4 & 22.1 & 81.38 & 2, 4 & 77.84, 81.11 \\
     & & ViT-B-8/4 & 86.6 & 84.54 & 2, 4 & 80.34, 83.98 \\
     & & ViT-L-8/4 & 304.3 & 85.84 & 2, 4 & 83.73, 85.45 \\
     & & EVA02-T-8/4 & 5.8 & 80.63 & 2, 4 & 66.32, 79.56 \\
     & & EVA02-S-8/4 & 22.1 & 85.73 & 2, 4 & 71.37, 84.70 \\
     & & EVA02-B-8/4 & 87.1 & 88.69 & 2, 4 & 84.62, 88.19 \\
     & & EVA02-L-8/4 & 305.1 & 90.05 & 2, 4 & 88.25, 89.72 \\
    \midrule
    \multirow{13}{*}{\textbf{TPC (Ours)}} & \multirow{6}{*}{CNN-to-SNN} & ResNet18-1/1 & 11.7 & 71.49 & 32, 64 & \cellcolor{gray!20}\textbf{70.12, 71.36} \\
    & & ResNet34-1/1 & 21.8 & 76.42 & 32, 64 & \cellcolor{gray!20}\textbf{60.94, 75.12} \\
     & & VGG-1/1 & 138 & 73.25 & 32, 64 & \cellcolor{gray!20}\textbf{72.36, 73.19} \\
     & & ResNet18-4/1 & 11.7 & 71.49 & 4, 8 & \cellcolor{gray!20}\textbf{71.03, 71.39} \\
     & & ResNet34-4/1 & 21.8 & 76.42 & 4, 8 & \cellcolor{gray!20}\textbf{73.89, 76.28} \\
     & & VGG-4/1 & 138 & 73.25 & 4, 8 & \cellcolor{gray!20}\textbf{72.85, 73.22} \\
     \cmidrule{2-7}
     & \multirow{7}{*}{Transformer-to-SNN} & ViT-S-8/4 & 22.1 & 81.38 & 2, 4 & \cellcolor{gray!20}\textbf{78.39, 81.26} \\
     & & ViT-B-8/4 & 86.6 & 84.54 & 2, 4 & \cellcolor{gray!20}\textbf{80.84, 84.33} \\
     & & ViT-L-8/4 & 304.3 & 85.84 & 2, 4 & \cellcolor{gray!20}\textbf{84.70, 85.62} \\
     & & EVA02-T-8/4 & 5.8 & 80.63 & 2, 4 & \cellcolor{gray!20}\textbf{69.52, 79.68} \\
     & & EVA02-S-8/4 & 22.1 & 85.73 & 2, 4 & \cellcolor{gray!20}\textbf{76.42, 85.13} \\
     & & EVA02-B-8/4 & 87.1 & 88.69 & 2, 4 & \cellcolor{gray!20}\textbf{86.29, 88.42} \\
     & & EVA02-L-8/4 & 305.1 & 90.05 & 2, 4 & \cellcolor{gray!20}\textbf{88.42, 89.83}\\
    \bottomrule
  \end{tabular}}
  \vspace{-2mm}
\end{table*}

\begin{table*}[htbp]
\centering
\caption{Comparing TPC with SOTA models of BERT on the GLUE evaluation set. * denotes non-convergence. \(^\dagger\) indicates additional time-steps required to gather the necessary prior information.}
\vspace{-2mm}
\label{tab:glue}
\resizebox{\textwidth}{!}{
\begin{tabular}{l c c c c c c c c c}
\toprule
\textbf{Model} & \textbf{Type} & \(T\) & \textbf{QQP} & \textbf{MNLI-m} & \textbf{SST-2} & \textbf{QNLI} & \textbf{RTE} & \textbf{MRPC} & \textbf{STS-B} \\
\midrule
BERT \cite{devlin2019bert} & ANN & / & 90.71 & 84.11 & 92.66 & 90.99 & 64.98 & 84.56/89.19 & 88.70/88.48 \\
CBoW \cite{wang2018glue} & ANN & / & 75.00 & 57.10 & 79.50 & 62.50 & 71.90 & 75.00/83.70 & 70.60/71.10 \\
GenSen \cite{subramanian2018learning} & ANN & / & 82.60 & 71.40 & 87.20 & 62.50 & 78.40 & 80.40/86.20 & 81.30/81.80 \\
\midrule
SNN-TextCNN \cite{lvspiking} & SNN & 50  & 0.00$^*$ & 64.91 & 80.91 & 64.91 & 47.29 & -/80.62     & 0.00$^*$/-  \\
Spikebert \cite{lv2025spikebert} & SNN & 4   & 68.17 & 71.42 & 85.39 & 66.37 & 57.47 & -/81.98     & -/18.73$^*$ \\
SpikeLM \cite{xing2024spikelm} & SNN & 4   & -     & 77.10 & 87.00 & 85.30 & 69.00 & -/85.70     & 84.90/-     \\
SpikingBERT \cite{bal2024spikingbert} & SNN & 60  & 86.82 & 78.10 & 88.19 & 85.20 & 66.06 & 79.17/85.15 & 82.20/81.90 \\
\midrule
SPR (BERT) \cite{hao2023reducing} & ANN-to-SNN & 8 (16$^\dagger$) & 87.48 & 77.56 & 90.48 & 87.75 & 64.98 & 78.68/85.76 & 86.71/86.50 \\
QCFS (BERT) \cite{buoptimal} & ANN-to-SNN & 8   & 88.42 & 79.57 & 89.91 & 86.80 & 56.68 & 78.92/85.37 & 86.18/85.82 \\
COS (BERT) \cite{haobridging} & ANN-to-SNN & 8 (8$^\dagger$)  & 88.85 & 79.91 & 89.79 & 87.37 & 63.18 & 79.66/86.33 & 86.49/86.23 \\
FAS (BERT) \cite{chen2025fas} & ANN-to-SNN & 4   & 90.38 & 82.77 & 91.17 & 90.13 & 66.06 & 86.02/90.22 & 87.46/87.26 \\
LAS (BERT) \cite{chen2026loss} & ANN-to-SNN & 16  & 90.73 & 84.19 & 92.66 & 90.92 & 65.34 & 84.80/89.42 & 88.76/88.53 \\
\midrule
\textbf{TPC (BERT)} & ANN-to-SNN & 4 & \cellcolor{gray!20}\textbf{90.90} & \cellcolor{gray!20}\textbf{84.07} & \cellcolor{gray!20}\textbf{92.43} & \cellcolor{gray!20}\textbf{90.62} & \cellcolor{gray!20}\textbf{64.56} & \cellcolor{gray!20}\textbf{84.75/89.55} & \cellcolor{gray!20}\textbf{87.84/88.12} \\
\textbf{TPC (BERT)} & ANN-to-SNN & 8 & \cellcolor{gray!20}\textbf{90.93} & \cellcolor{gray!20}\textbf{84.15} & \cellcolor{gray!20}\textbf{92.68} & \cellcolor{gray!20}\textbf{91.54} & \cellcolor{gray!20}\textbf{65.41} & \cellcolor{gray!20}\textbf{84.93/89.68} & \cellcolor{gray!20}\textbf{88.79/88.56} \\
\bottomrule
\end{tabular}}
\vspace{-2mm}
\end{table*}

\begin{table*}[htbp]
    \centering
    \caption{Performance comparison of different methods across various VLM benchmarks}
    \vspace{-2mm}
    \label{tab:vlm}
    \resizebox{\textwidth}{!}{
    \begin{tabular}{lcccccccc}
        \toprule
        \textbf{Method} & \(T\) & \textbf{ScienceQA \cite{lu2022learn}} & \textbf{RealWorldQA \cite{xai2024grok15v}} & \textbf{BLINK \cite{fu2024blink}} & \textbf{POPE \cite{li2023evaluating}} & \textbf{HallusionBench \cite{guan2024hallusionbench}} & \textbf{MMStar \cite{chen2024we}} & \textbf{MME \cite{fu2023mme}} \\ 
        \midrule
        LLaVA-1.5-7B \cite{lin2026text} & / & 67.22 & 52.41 & 41.22 & 82.81 & 46.90 & 33.40 & 1732.41 \\
        Qwen2-VL-7B \cite{wang2024qwen2} & / & 85.09 & 67.18 & 51.55 & 86.32 & 50.60 & 60.70 & 2326.80 \\
        Qwen2.5-VL-7B \cite{bai2025qwen25vltechnicalreport} & / & 82.53 & 68.50 & 47.60 & 86.30 & 52.90 & 63.90   & 2347.03 \\
        \cmidrule{1-9}
        LAS \cite{chen2026loss} (LLaVA-1.5-7B) & 16 & 66.38 & 49.15 & 41.01 & 80.79 & 43.12 & 31.86 & 1613.04 \\
        LAS \cite{chen2026loss} (Qwen2-VL-7B) & 16 & 81.40 & 66.79 & 52.18 & 84.77 & 49.32 & 53.86 & 2222.59 \\
        \cmidrule{1-9}
        \textbf{TPC (LLaVA-1.5-7B)} & 4 & \cellcolor{gray!20}\textbf{66.49} & \cellcolor{gray!20}\textbf{53.86} & \cellcolor{gray!20}\textbf{41.13} & \cellcolor{gray!20}\textbf{83.04} & \cellcolor{gray!20}\textbf{45.74} & \cellcolor{gray!20}\textbf{32.80} & \cellcolor{gray!20}\textbf{1729.82} \\
        \textbf{TPC (Qwen2-VL-7B)} & 4 & \cellcolor{gray!20}\textbf{83.60} & \cellcolor{gray!20}\textbf{67.15} & \cellcolor{gray!20}\textbf{52.27} & \cellcolor{gray!20}\textbf{85.62} & \cellcolor{gray!20}\textbf{50.63} & \cellcolor{gray!20}\textbf{54.60} & \cellcolor{gray!20}\textbf{2232.29} \\
        \textbf{TPC (Qwen2.5-VL-7B)} & 4 & \cellcolor{gray!20}\textbf{80.57} & \cellcolor{gray!20}\textbf{68.06} & \cellcolor{gray!20}\textbf{45.29} & \cellcolor{gray!20}\textbf{84.86} & \cellcolor{gray!20}\textbf{51.26} & \cellcolor{gray!20}\textbf{60.80} & \cellcolor{gray!20}\textbf{2257.13} \\
        \bottomrule
    \end{tabular}}
    \vspace{-2mm}
\end{table*}

\subsection{Expected Quantization Noise Minimization and Damping Control}

Within the TPC framework, although macro-level expectation tracking ensures lossless feature reconstruction, the ideal prediction error (i.e., \(\tilde{x}^l[t]-\mathcal{P}_t\)) computed at the transmitter is ultimately compressed by a multi-threshold quantization function \(\mathcal{Q}(\cdot)\) into ultra-low bit-width discrete spikes for asynchronous transmission. Consequently, long-tail outliers within the model, coupled with low time-steps (e.g., \(T\le4\)), induce severe truncation and rounding errors. To minimize the expected quantization noise introduced during the single-step quantization process, we conduct a theoretical analysis of the optimal decision boundaries for \(\mathcal{Q}(\cdot)\).


\begin{theorem}
\label{thm: expected quantization noise minimization and damping control}
Let the candidate discrete spike set for a neuron in the \(l\)-th layer be a strictly symmetric exponentially decaying set \(\Lambda^l = \{\lambda^l_1, \dots, \lambda^l_n\} \cup \{-\lambda^l_1, \dots, -\lambda^l_n\} \cup \{0\}\), where adjacent thresholds of the same polarity satisfy \(\lambda^l_{i-1} = 2\lambda^l_i\). Under the assumption of locally uniformly distributed residuals, to minimize the mean squared quantization error \(E = \mathbb{E}\big[(V - \mathcal{Q}(V))^2\big]\) between the input membrane potential \(V\) and the actual emitted spike \(x\), the analytically optimal decision boundary of the quantization function \(\mathcal{Q}(\cdot)\) must be defined as:
\begin{equation}
    \mathcal{Q}(V) = \text{sgn}(V) \cdot \lambda^l_i, \quad \text{if } |V| \in [\beta^* \lambda^l_i, \beta^* \lambda^l_{i-1}),
\end{equation}
where the theoretical optimal decision offset coefficient is \(\beta^* \equiv 0.75\) (proof in \cref{supp: proof of the theorem 3.6.}).
\end{theorem}

Although the \cref{thm: expected quantization noise minimization and damping control} minimizes single-step quantization noise, historical quantization errors retained within the membrane potential \(V(t)\) may trigger temporal oscillations and signal overshoots in exceedingly deep network layers. To address this, a prior damping coefficient \(\alpha \in [0, 1)\) is introduced into the TPC dynamics to facilitate momentum decay (refer to \cref{equ: dampling control}). This effectively prevents neurons from overreacting to transient errors, ensuring that the model rapidly converges to a steady state at low time-steps.

\section{Experiments}

\subsection{Comparative Study}

To comprehensively evaluate the effectiveness of TPC in ANN-to-SNN conversion, we assess it across vision, language, and multimodal tasks. The evaluated architectures include CNNs, Transformers, BERT and Vision-Language Models (VLMs). Additional experiment results see \cref{supp: additional experiment results}.

\textbf{Results on ImageNet-1K.} We conduct conversion experiments on 10 different CNNs and Transformers using the ImageNet-1K dataset (\cref{tab:imagenet}). Following \cite{huang2025differential}, we denote the converted model as \(model-n/c\). In this notation, the multi-threshold neuron has \(n\) positive and \(n\) corresponding negative thresholds, and the computed channel thresholds are scaled by a factor of \(c\). When \(n=1\), it acts as an IF neuron with an extra negative threshold. In CNNs, when \(n=1\), our approach outperforms existing methods on identical architectures, achieving state-of-the-art results. For \(n>1\), we obtain superior performance using shorter time-steps. In Transformers, we scale the statistical thresholds by setting \(n=8\) and \(c=4\). Our training-free method achieves high conversion performance within shorter time-steps.

\textbf{Results on GLUE Benchmark.} TPC achieves nearly lossless conversion in language understanding tasks. As shown in \cref{tab:glue}, with \(T=4\), TPC reaches accuracies highly comparable to BERT across multiple tasks, including SST-2 and QNLI. On the QQP task, TPC even outperforms BERT by \(0.19\%\). Furthermore, TPC surpasses existing SNN methods while requiring fewer time-steps.

\textbf{Results on VLM Benchmarks.} As shown in \cref{tab:vlm}, TPC shows minimal performance degradation during VLM conversion. Applied to Qwen2-VL-7B, TPC scores 83.60 on ScienceQA and 67.15 on RealWorldQA, closely matching the ANN baselines. Notably, TPC even slightly outperforms the ANN baselines on BLINK (Qwen2-VL-7B) and POPE (LlaVA-1.5-7B). Compared to prior methods, TPC delivers higher conversion performance using fewer time-steps.

\subsection{Energy Estimation and Result Analysis}

\begin{figure*}[htbp]
    \centering
    \includegraphics[width=\linewidth]{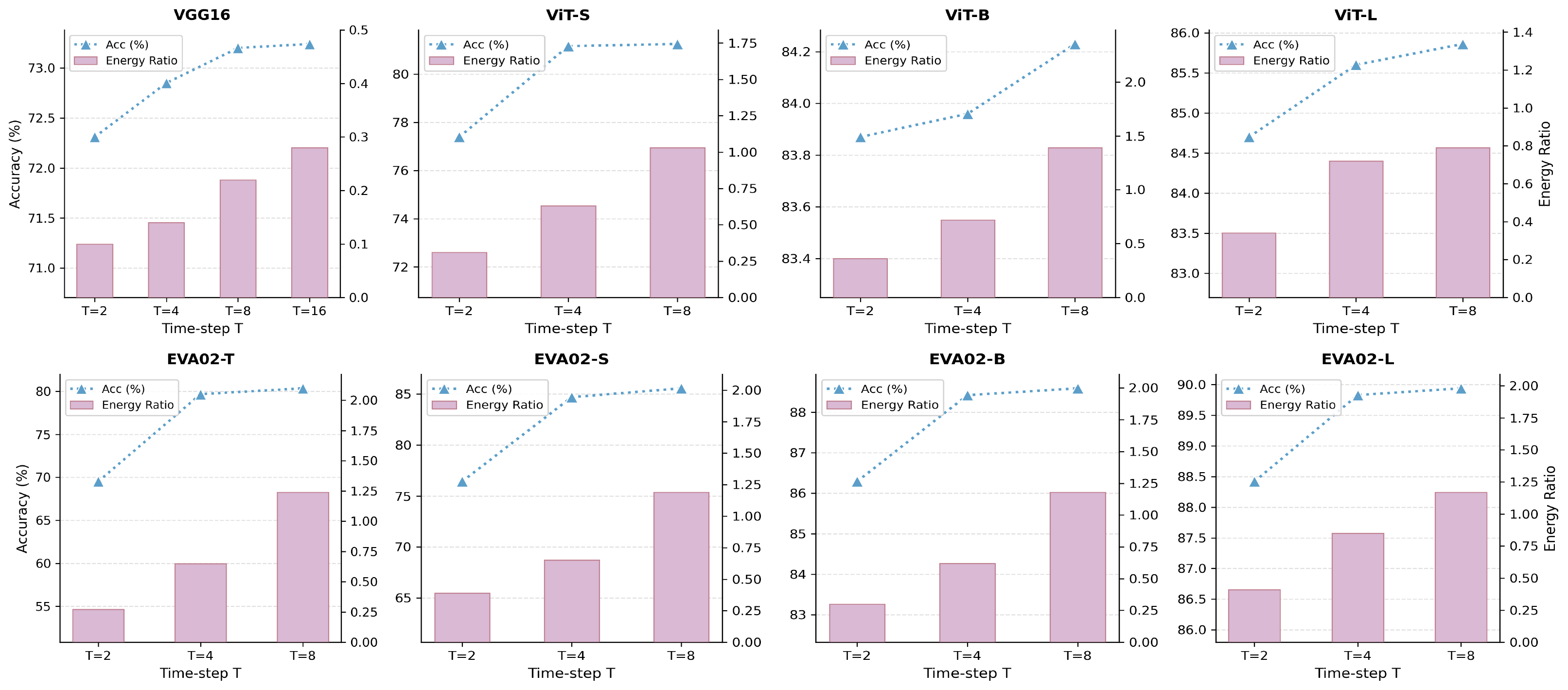}  
    \vspace{-6mm}
    \caption{Relative energy consumption of the converted SNN compared to the ANN.}
    \vspace{-3mm}
    \label{fig:energy ratio}
\end{figure*}

\begin{figure*}[htbp]
    \centering
    \includegraphics[width=\linewidth]{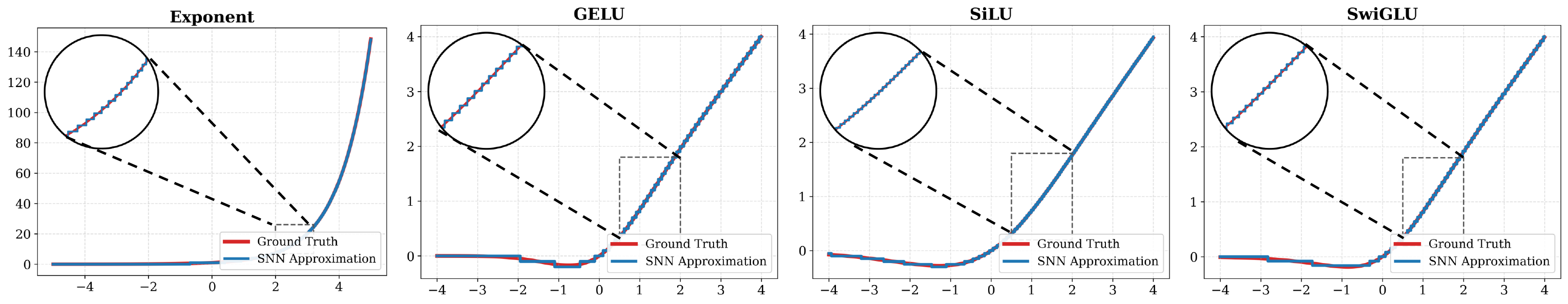}  
    \vspace{-6mm}
    \caption{Illustrations of nonlinear function approximations.}
    \vspace{-4mm}
    \label{fig:efficacy studies}
\end{figure*}

Based on \cite{horowitz20141}, we evaluate the relative energy consumption of the converted SNN compared to the ANN (evaluate on ImageNet-1K, as shown in \cref{fig:energy ratio}) using \cref{equ: energy estimation}, where \(E_{\text{MAC}}\)=4.6pJ and \(E_{\text{AC}}\)=0.9pJ.
\begin{equation}
\label{equ: energy estimation}
    \frac{E_{\text{SNN}}}{E_{\text{ANN}}}=\frac{\text{MACs}_{\text{SNN}}*E_{\text{MAC}}+\text{ACs}_{\text{SNN}}*E_{\text{AC}}}{\text{MACs}_{\text{ANN}}*E_{\text{MAC}}}.
\end{equation}
Most network computations occur in the fully connected, convolutional, and matrix multiplication layers. Sinces SNNs implement these primarily through addition operations (\(\text{ACs}_{\text{SNN}} \gg \text{MACs}_{\text{SNN}}\)), we approximate \(\text{MACs}_{\text{SNN}} \approx 0\) \cite{huang2025differential}. Consequently, we use the firing rate to estimate \(\frac{\text{ACs}_{\text{SNN}}}{\text{MACs}_{\text{ANN}}}\), which allows us to determine the overall energy consumption ratio. Our method achieves SNN performance comparable to that of ANNs with low power consumption and extremely short time-steps.

\begin{figure}[t]
    \centering
    \includegraphics[width=\linewidth]{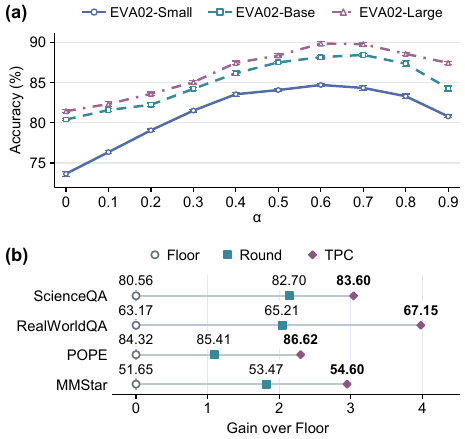}  
    \vspace{-8mm}
    \caption{(a) Impact of damping control coefficient; (b) Validation of optimal decision boundary.}
    \vspace{-6mm}
    \label{fig:ablation alpha and decision boundary}
\end{figure}

\subsection{Ablation Study}

\textbf{Impact of Damping Control Coefficient \(\alpha\).} To suppress temporal oscillations in deep networks, we introduce a damping control coefficient, \(\alpha\), into the PreSN dynamics. \cref{fig:ablation alpha and decision boundary}(a) illustrates the conversion accuracy on EVA02 under varying \(\alpha\) values. Results show that as \(\alpha \to 0\), the model overreacts to single-step prediction errors, causing a sharp drop in accuracy. For \(\alpha \in [0.5, 0.75]\), the damping control effectively improves conversion accuracy, allowing performance to peak. Conversely, when \(\alpha > 0.8\), excessive damping makes the PreSNs sluggish in responding to prediction errors, leading to performance degradation.

\textbf{Validation of Optimal Decision Boundary \(\beta^*\).} Based on \cref{thm: expected quantization noise minimization and damping control}, to minimize the expected single-step quantization noise in the presence of long-tail outliers, the analytically optimal decision boundary should be set to the average of adjacent thresholds, \(\beta^*\). To validate this derivation, we substitute the TPC quantization function with Floor (\(\beta=1.0\)) and Round (\(\beta=0.5\)) operations, and evaluate the variants on Qwen2-VL-7B. Experimental results (\cref{fig:ablation alpha and decision boundary}(b)) indicate that employing either Floor or Round leads to varying degrees of performance degradation across all four tasks.

\subsection{Approximation Capabilities Study}


We evaluate how faithfully PreSNs approximate GELU, SiLU, SwiGLU, and the exponential function within eight time steps. These functions span smooth activations, multiplicative gating, and exponential growth, allowing us to assess approximation fidelity across different curve shapes and rates of change. As shown in \cref{fig:efficacy studies}, the PreSN outputs nearly overlap with the reference curves, capturing both the transition regions of the activation functions and the steep rise of the exponential function. The close agreement across all four functions shows that PreSNs can represent diverse nonlinear transformations within a short temporal window.

\section{Conclusion}

In this work, we propose the Temporal Predictive Coding (TPC) framework for ANN-to-SNN conversion. TPC leverages temporal dynamics to predict current neuronal states and treats discrete spikes as corrections to prediction errors. Furthermore, we theoretically analyze the expected error introduced during conversion and derive the analytically optimal decision boundary for multi-threshold quantization. To address the accumulation of residual errors in deep networks, we introduce a damping control mechanism that ensures rapid convergence under short time-steps. Extensive experiments across vision (CNNs and Transformers), language (BERT) and multimodal (VLMs) tasks demonstrate that the proposed TPC framework achieves state-of-the-art performance while minimizing both time-steps and energy consumption. Despite these advances, the current TPC framework is primarily validated on dense architectures. Future research will focus on addressing this limitation by extending TPC to accommodate sparse, MoE-based Large Language Models (LLMs) and scaling it to even larger architectures (see \cref{supp: limitations} for a detailed discussion).

\section*{LLM Usage Statement}

The manuscript was drafted entirely by the authors, with LLMs (GPT-6.1 sol) utilized strictly as a post-writing editorial tool to refine sentence structure, clarity, and academic tone. The model was not involved in literature synthesis, experimental design, or initial text generation, and all suggested edits were manually verified by the authors to ensure technical accuracy and academic integrity.

{
    \small
    \bibliographystyle{ieeenat_fullname}
    \bibliography{main}
}

\clearpage
\renewcommand{\thesection}{\Alph{section}}
\setcounter{section}{0}
\setcounter{page}{1}
\maketitlesupplementary

\section{Supplementary Material}
\subsection{Discussion}
\label{supp: discussion}

In section \ref{section: methodology}, we introduce an ANN-to-SNN conversion framework based on Temporal Predictive Coding (TPC). While directly inspired by the efficient predictive coding mechanisms of the cerebral cortex, the "predict-update" dynamics of TPC neurons demonstrate a profound mathematical isomorphism with the discrete-time Kalman Filter (KF) \cite{kalman1960new,welch1995introduction} from the perspective of modern control theory and statistical signal processing. Rather than merely approximating activation values, the TPC framework constructs a recursive state estimator to optimally track the ideal continuous activations of the source ANN. Specifically, the core dynamic mapping between the two can be summarized across the following four dimensions:

\textbf{Isomorphism in Predict-Update Dynamics.} The core of the Kalman Filter (KF) lies in approximating the true state through a \textbf{priori estimation (Predict)} and a \textbf{posteriori updating (Update)}. In the TPC framework, the operational logic of PreSNs perfectly aligns with this paradigm:
\begin{itemize}
    \item \textbf{Time Update (Prior Prediction)}: In the KF, the system performs an a priori prediction based on a state transition matrix, formulated as \(\hat{x}_{k|k-1} = F \hat{x}_{k-1|k-1}\). In TPC, this directly corresponds to extracting temporal inertia using historical momentum: \(\mathcal{P}_t = \alpha \cdot \hat{x}^l[t-1]\). Here, the damping coefficient \(\alpha\) serves the role of the state transition matrix.
    \item \textbf{Measurement Innovation}: In the KF, the system calculates the residual (innovation) between the actual observation and the prior prediction. At the TPC transmitting end, the discrete spike residual computed and emitted by the neuron, \(x^l[t] = \mathcal{Q}\big( \tilde{x}^l[t] - \mathcal{P}_t \big)\), precisely represents the system innovation carrying quantization noise. This demonstrates that TPC transmits pure "unpredicted information" across layers, rather than redundant states.
    \item \textbf{Measurement Update (Posterior Reconstruction)}: In the KF, the system performs an a posteriori state update by combining the innovation with the Kalman gain: \(\hat{x}_{k|k} = \hat{x}_{k|k-1} + K_k \tilde{y}_k\). At the TPC receiving end, this process is elegantly and efficiently reduced to a simple addition: \(\hat{x}^l[t] = \mathcal{P}_t + x^l[t]\). In essence, newly arrived spikes are utilized to correct the prior prediction.
\end{itemize}

\textbf{Steady-State Tracking and Recursive Smoothing.} In the KF, the output converges to the true state with minimum mean square error as time progresses. The expected tracking equation defined in TPC, \(r^l[t] = \frac{t-1}{t}r^l[t-1] + \frac{\hat{x}^l[t]}{t}\), essentially functions as an incremental Monte Carlo integration (or a recursive averaging filter). Because we rigorously prove the expected equivalence of different operators (\cref{thm: tpc for single-input nonlinear layers} and \cref{thm: tpc for multi-input nonlinear layers}), the internal tracking variable \(r^l[t]\) acts as an unbiased estimate of the system's true continuous state. As discrete spike sequences continuously arrive, \(r^l[t]\) converges rapidly, completely eliminating the exponential decay error inherent in rate coding.

\textbf{Damping Control.} In the KF, the magnitude of the gain \(K_k\) depends on the ratio of the prediction error covariance to the observation noise covariance. In ANN-to-SNN conversion, the primary noise stems from truncation and rounding errors caused by extremely low bit-widths (i.e., "observation noise"). Furthermore, because the quantization error from the previous step is stored in the membrane potential and propagated to the next step ("process noise" induced by error accumulation), the system is highly susceptible to temporal oscillations in deep networks. To address this, TPC introduces a prior damping coefficient \(\alpha \in [0, 1)\) to replace the complex dynamic gain matrix. \(\alpha\) restricts the system's over-reliance on historical states, effectively introducing a pole decay. This prevents the neuron from overreacting to transient quantization errors. Under extremely low time-steps (e.g., \(T \le 4\)), this mechanism stabilizes the system in a manner similar to the Kalman gain.

\textbf{Quantization Noise Minimization vs. MMSE.} The Kalman Filter is the proven Minimum Mean Square Error (MMSE) optimal estimator for linear Gaussian systems. While SNN spike generation is a highly non-linear discrete process where quantization noise is typically non-Gaussian, we theoretically analyze the quantization function \(\mathcal{Q}(\cdot)\) in \cref{thm: expected quantization noise minimization and damping control}. Assuming a locally uniform distribution, we rigorously minimize the mean square quantization error, \(E = \mathbb{E}\big[(V - \mathcal{Q}(V))^2\big]\), to derive the theoretically optimal decision offset coefficient (\(\beta^* \equiv 0.75\)). This demonstrates that, even at the level of local discrete spike generation, TPC adheres to the same optimal state estimation philosophy of minimizing the mean square error.

\subsection{Preliminary: Multi-threshold Spiking Neuron}
\label{supp: multi-threshold spiking neuron}

Many previous studies have used ternary-valued neurons to simulate negative values and reduce conversion errors \cite{li2022quantization,wang2022signed,you2024spikezip}. Furthermore, \cite{huang2024towards} introduced a multi-channel approach to implement multi-threshold (MT) spiking neurons. In this paper, we also adopt spiking MT neurons. The MT neuron is characterized by the following parameters: the base threshold $\theta$ and the total number of thresholds \(2n\) (where \(n\) are positive thresholds and \(n\) are negative thresholds). The thresholds of an MT neuron are indexed by \(i\), where \(\lambda^l_i\) denotes the \(i\)-th threshold in layer \(l\):
\begin{align}
\label{equ: thresholds of MT neurons}
& \lambda^l_1=\theta^l, \lambda^l_2=\frac{\theta^l}{2},...,\lambda^l_n=\frac{\theta^l}{2^{n-1}}, \\
& \lambda^l_{n+1}=-\theta^l,\lambda^l_{n+2}=-\frac{-\theta^l}{2},...,\lambda^l_{2n}=\frac{-\theta^l}{2^{n-1}}.
\end{align}

Let the variables \(I^l[t], W^l, s^l_i[t], x^l[t], m^l[t]\), and \(v^l[t]\) denote the input current, weights, the output spike for the \(i\)-th threshold, the total output signal, and the membrane potential before and after the spike in layer \(l\) at time step \(t\), respectively. The dynamics of an MT neuron are described by the following equations:
\begin{align}
& m^l[t]=v^l[t-1]+I^l[t]=v^l[t-1]+x^{l-1}[t], \\
& s^l_i[t]=\text{MTH}_{\theta,n}(m^l[t],i), \\
& x^l[t]=\sum_i s^l_i[t]W^l \lambda^l_i, \\
& v^l[t]=m^l[t]-x^l[t], \\
& \text{MTH}_{\theta,n}(m^l[t],i)=
\begin{cases}
0, \ \text{if} \ \lambda_{2n}<x<\lambda_n \\
1, \ \text{elif} \ i=\text{argmin}_p|x-\lambda_p| \\
0, \ \text{else}
\end{cases}.
\end{align}
Fig.\ref{fig:neuron and predictive coding}(a) illustrates the dynamics of MT neurons; when \(n=1\), the model simplifies to an IF neuron with an additional negative threshold. 

All experiments were conducted on eight NVIDIA RTX 4090 GPUs.


\subsection{Main Theorems and Proofs}
\label{suppl: main theorems and proofs}

\subsubsection{Proof of Definition \ref{def: temporal predictive coding}}
\label{supp: proof of the definition 4.1.}

\begin{customdef}{3.1}
In temporal predictive coding (TPC), denote \(x^l[t]\) as the actual transmitted prediction error of the neuron at time-step \(t\). Define \(\hat{x}^l[t]\) as the reconstructed raw differential output, and \(r^l[t]\) as the encoded continuous activation value at time-step \(t\). The relationship among them is expressed as follows:
\begin{align}
    \label{equ: xl[t]}
    & \hat{x}^l[t]=x^l[t]+\mathcal{P}(\hat{x}^l[t-1]), \\
    & r^l[t]=\frac{t-1}{t} r^l[t-1]+\frac{\hat{x}^l[t]}{t},
    \label{equ: r[t]}
\end{align}
where \(\mathcal{P}(\cdot)\) represents the damped temporal predictive function based on historical momentum, \(t\) starts from \(1\), and \(r^l[0]=0\).
\end{customdef}

\begin{proof}

Equ. \ref{equ: r[t]} implements an online arithmetic mean of the reconstructed outputs. At the first time-step, substituting \(t=1\) into Equation \ref{equ: r[t]} gives
\begin{equation}
    r^l[1]=\hat{x}^l[1].
\end{equation}

For an arbitrary time-step \(i\geq 1\), multiplying Equation (25) by \(i\) yields
\begin{equation}
    i\cdot r^l[i]=(i-1)\cdot r^l[i-1]+\hat{x}^l[i].
\end{equation}

Equivalently, the reconstructed output at time-step \(i\) can be written as the temporal difference between two consecutive cumulative representations:
\begin{equation}
    \hat{x}^l[i]=i\cdot r^l[i]-(i-1)\cdot r^l[i-1].
\end{equation}

Summing both sides from \(i=1\) to \(t\), we obtain
\begin{equation}
    \sum_{i=1}^{t}\hat{x}^l[i]=\sum_{i=1}^{t}\left(i\cdot r^l[i]-(i-1)\cdot r^l[i-1]\right).
\end{equation}

Expanding the right-hand side gives
\begin{align}
    \sum_{i=1}^{t}\hat{x}^l[i]
={}&
\left(r^l[1]-0\cdot r^l[0]\right)
+\left(2r^l[2]-r^l[1]\right)
+\cdots \notag \\
&+
\left(tr^l[t]-(t-1)r^l[t-1]\right).
\end{align}

The intermediate terms cancel pairwise, forming a telescoping sum. Therefore,
\begin{equation}
    \sum_{i=1}^{t}\hat{x}^l[i]=t\cdot r^l[t]-0\cdot r^l[0]=t\cdot r^l[t],
\end{equation}
where the last equality follows from the initial condition \(r^l[0]=0\). Dividing both sides by \(t\) gives the explicit form of the encoded continuous activation:
\begin{equation}
    \label{equ: equ32}
    r^l[t]=\frac{1}{t}\sum_{i=1}^{t}\hat{x}^l[i].
\end{equation}

Thus, Equ. \ref{equ: r[t]} is exactly the recursive implementation of the finite-time arithmetic mean of all reconstructed outputs. Substituting the reconstruction rule in Equ. \ref{equ: xl[t]} into Equ. \ref{equ: equ32} yields
\begin{equation}
    r^l[t]=\frac{1}{t}\sum_{i=1}^{t}\left(x^l[i]+\mathcal{P}\!\left(\hat{x}^l[i-1]\right)\right).
\end{equation}

Equivalently, the encoded activation can be decomposed as
\begin{equation}
    r^l[t]=\frac{1}{t}\sum_{i=1}^{t}x^l[i]+\frac{1}{t}\sum_{i=1}^{t}\mathcal{P}\!\left(\hat{x}^l[i-1]\right).
\end{equation}

This derivation establishes that \(r^l[t]\) is the exact temporal average of the reconstructed outputs \(\hat{x}^l[i]\). Each reconstructed output consists of two components: the sparse prediction error \(x^l[i]\), which is physically transmitted between layers, and the predictive compensation \(\mathcal{P}(\hat{x}^l[i-1])\), which is generated locally from the historical state. Consequently, TPC reconstructs the continuous activation while communicating only the unpredictable residual errors. Moreover, the factor \((t-1)/t\) in Equation (25) preserves the complete cumulative contribution of the previous \(t-1\) reconstructed outputs without repeatedly adding or double-counting the historical representation.

\end{proof}

\subsubsection{Proof of \cref{thm: neurons in rate coding to presns}}
\label{supp: proof of the theorem 4.2.}

\begin{customthm}{3.2}
In rate coding, the output pulse \(s^{l-1}[t]\) from the previous layer generates the input current to the current layer via the synaptic weight \(W^l\), i.e., \(I^l[t] = W^l s^{l-1}[t]\). Within the TPC framework, by introducing a local state \(m^l[t]\), any spiking neuron can be equivalently converted into a PreSN. The dynamical update equation for its equivalent membrane current \(I^l[t]\) is as follows:
\end{customthm}

\begin{proof}

According to \cref{thm: tpc for linear layers}, the static bias of the source ANN is converted into a one-time input applied at the first time-step. Therefore, the ideal target signal of layer \(l\) can be written as
\begin{equation}
    \tilde{x}^l[t]=W^l\hat{x}^{l-1}[t]+b^l\delta_{t1},
\end{equation}
where \(\delta_{t1}=1\) if \(t=1\) and \(\delta_{t1}=0\) otherwise. To ensure that only prediction errors are transmitted between layers, the effective input current of the PreSN is defined as the difference between the ideal target signal and the current-layer prior prediction:
\begin{equation}
    \label{equ: equ36}
    I^l[t]=\tilde{x}^l[t]-P_t^l, \qquad P_t^l=\alpha\hat{x}^l[t-1].
\end{equation}

We first consider the initial time-step \(t=1\). Under the resting-state initialization
\begin{equation}
    \label{equ: equ37}
    \hat{x}^{l-1}[0]=\hat{x}^l[0]=0,
\end{equation}
the prior predictions of both layers vanish, i.e., \(P_1^{l-1}=P_1^l=0\). Consequently, \(\hat{x}^{l-1}[1]=x^{l-1}[1]\), and Equa. \ref{equ: equ36} becomes
\begin{equation}
    I^l[1]=W^l x^{l-1}[1]+b^l.
\end{equation}

We next consider \(t>1\). Since the bias is injected only at the first time-step, \(\delta_{t1}=0\). Using the reconstruction rule
\begin{equation}
    \label{equ: equ39}
    \hat{x}^{l-1}[t]=P_t^{l-1}+x^{l-1}[t]=\alpha\hat{x}^{l-1}[t-1]+x^{l-1}[t],
\end{equation}
the ideal target signal can be expanded as
\begin{equation}
    \label{equ: equ40}
    \tilde{x}^l[t]=\alpha W^l\hat{x}^{l-1}[t-1]+W^l x^{l-1}[t].
\end{equation}

Substituting Equation (38) and \(P_t^l=\alpha\hat{x}^l[t-1]\) into Equ. \ref{equ: equ36} gives
\begin{equation}
    I^l[t]=W^l x^{l-1}[t]+\alpha\left(W^l\hat{x}^{l-1}[t-1]-\hat{x}^l[t-1]\right), \qquad t>1.
\end{equation}

We define the local tracking state as
\begin{equation}
\label{equ: local tracking state}
    m^l[t]=W^l\hat{x}^{l-1}[t-1]-\hat{x}^l[t-1], \qquad m^l[1]=0.
\end{equation}

This state measures the discrepancy between the bias-free linear projection of the previous layer’s posterior reconstruction and the current layer’s posterior reconstruction. Substituting Equ. \ref{equ: equ40} into Equ. \ref{equ: equ37} and Equ. \ref{equ: equ39} yields the piecewise current dynamics
\begin{equation}
    I^l[t]=
\begin{cases}
W^l x^{l-1}[t]+b^l, & t=1,\\[2mm]
W^l x^{l-1}[t]+\alpha m^l[t], & t>1.
\end{cases}
\end{equation}

It remains to derive the recursive update of \(m^l[t]\). Applying its definition at the next time-step gives
\begin{equation}
    m^l[t+1]=W^l\hat{x}^{l-1}[t]-\hat{x}^l[t].
\end{equation}

Using the reconstruction equations
\begin{align}
    & \hat{x}^{l-1}[t]=\alpha\hat{x}^{l-1}[t-1]+x^{l-1}[t], \\
    & \hat{x}^l[t]=\alpha\hat{x}^l[t-1]+x^l[t],
\end{align}

Equ. \ref{equ: local tracking state} can be expanded as
\begin{align}
m^l[t+1]
&= W^l\bigl(
    \alpha\hat{x}^{l-1}[t-1]+x^{l-1}[t]
    \bigr) \notag\\
&\quad{}-\bigl(
    \alpha\hat{x}^{l}[t-1]+x^{l}[t]
    \bigr) \notag\\
&= \alpha\bigl(
    W^l\hat{x}^{l-1}[t-1]-\hat{x}^{l}[t-1]
    \bigr) \notag\\
&\quad{}+W^l x^{l-1}[t]-x^{l}[t] \notag\\
&= \alpha m^l[t]+W^l x^{l-1}[t]-x^l[t].
\end{align}

Therefore, the ANN bias is injected exactly once at \(t=1\), while the local state \(m^l[t]\) recursively tracks the remaining reconstruction discrepancy for \(t>1\). This establishes both dynamical equations in \cref{thm: neurons in rate coding to presns}.

\end{proof}

\subsubsection{Proof of \cref{thm: tpc for linear layers}}
\label{supp: proof of the theorem 4.3.}

\begin{customthm}{3.3}
For linear layers, including fully connected layers and convolutional layers, which can be expressed as
\begin{equation}
    r^l = W^l r^{l-1} + b^l,
\end{equation}
where \(W^l\) and \(b^l\) represent the weights and biases of the \(l\)-th layer, respectively. Within the TPC framework, the conversion for a linear layer is equivalent to eliminating the static bias term \(b^l\) from the forward computation and incorporating the bias term into the initial state of the subsequent PreSN at the first time step:
\begin{equation}
\hat{x}^l[t] = 
\begin{cases} W^l \hat{x}^{l-1}[t] + b^l, & \text{if } t = 1 \\ 
W^l \hat{x}^{l-1}[t], & \text{if } t > 1 
\end{cases}
\end{equation}
\end{customthm}

\begin{proof}
In ANN, let the absolute target activation value for a given input be \(y^l\): \(y^l = Wy^(l-1) + b^l\). When this network is converted to an SNN, since the network is in a resting state at \(t = 0\), the static ANN bias \(b^l\) is equivalent in spatiotemporal dynamics to a constant step input that is activated at the time of the stimulus input (\(t = 1\)). Therefore, by introducing a discrete step function \(u[t]\) (where \(u[t] = 1\) for \(t \ge 1\) and \(u[t] = 0\) for \(t \le 0\)), the target cumulative state of layer \(l\) at any time \(t\) can be expressed as:
\begin{equation}
    y^l[t]=W^l y^{l-1}[t]+b^l \cdot u[t].
\end{equation}
In the TPC framework, the reconstructed output \(\hat{x}^l[t]\) physically represents the temporal difference of that absolute state. By definition, we have:
\begin{equation}
\label{equ: current layer}
    \hat{x}^l[t] = y^l[t] - y^l[t-1].
\end{equation}

\textbf{a. When \(t=1\)}: Substituting \(t=1\) and the initial stationary conditions \(u[0]=0\) and \(y^l[0]=0\):
\begin{align}
& \hat{x}^l[1] = y^l[1] - y^l[0], \\
& \hat{x}^l[1] = (W^l y^{l-1}[1] + b^l \cdot u[1]) - 0, \\
& \hat{x}^l[1] = W^l y^{l-1}[1] + b^l.
\end{align}

Since the stationary condition is also satisfied in the previous layer, we have \(y^{l-1}[1] - y^{l-1}[0] = \hat{x}^{l-1}[1]\), which implies \(y^{l-1}[1] = \hat{x}^{l-1}[1]\). Substituting this into the above equation yields:
\begin{equation}
    \hat{x}^l[1] = W^l \hat{x}^{l-1}[1] + b^l.
\end{equation}
Therefore, in the first time-step of the temporal expansion, the bias \(b^l\) is explicitly retained.

\textbf{b. When \(t>1\)}: According to the TPC definition, the absolute cumulative activation states \(y^{l-1}[t]\) and \(y^{l-1}[t-1]\) of the previous layer at time-step \(t\) and \(t-1\), respectively, can be strictly expanded as a series sum of the historical reconstruction signal:
\begin{align}
y^{l-1}[t]
&= \sum_{i=1}^{t}\hat{x}^{l-1}[i] \notag\\
&= \hat{x}^{l-1}[1]+\hat{x}^{l-1}[2]+\cdots \notag\\
&\quad{}+\hat{x}^{l-1}[t-1]+\hat{x}^{l-1}[t],
\label{equ: series}\\
y^{l-1}[t-1]
&= \sum_{i=1}^{t-1}\hat{x}^{l-1}[i] \notag\\
&= \hat{x}^{l-1}[1]+\hat{x}^{l-1}[2]+\cdots \notag\\
&\quad{}+\hat{x}^{l-1}[t-1].
\end{align}

When \(t > 1\), \(u[t] = 1\) and \(u[t-1] = 1\). We substitute the target state of the current layer \(y^l\) into Equ.\ref{equ: current layer}:
\begin{equation}
\label{equ: expand}
    \hat{x}^l[t] = \Big( W^l y^{l-1}[t] + b^l \Big) - \Big( W^l y^{l-1}[t-1] + b^l \Big). 
\end{equation}

Expand the series in Equ.\ref{equ: series} and substitute it into Equ.\ref{equ: expand} to perform a linear expansion:
\begin{equation}
\begin{aligned}
\hat{x}^l[t]
&= \Biggl[
W^l\Biggl(
\sum_{i=1}^{t}\hat{x}^{l-1}[i]
\Biggr)+b^l
\Biggr]\\
&\quad{}-\Biggl[
W^l\Biggl(
\sum_{i=1}^{t-1}\hat{x}^{l-1}[i]
\Biggr)+b^l
\Biggr].
\end{aligned}
\end{equation}

By applying the linear distribution law of matrix multiplication, we multiply the weight \(W^l\) into the series and rearrange the polynomial by pairing terms corresponding to the same time step, yielding:
\begin{align} 
\hat{x}^l[t] =\ & W^l \big( \hat{x}^{l-1}[1] - \hat{x}^{l-1}[1] \big) \notag \\ 
&+ W^l \big( \hat{x}^{l-1}[2] - \hat{x}^{l-1}[2] \big) \notag \\
&+ \dots \notag \\
&+ W^l \big( \hat{x}^{l-1}[t-1] - \hat{x}^{l-1}[t-1] \big) \notag \\ 
&+ W^l \hat{x}^{l-1}[t] \notag \\
&+ \big( b^l - b^l \big). 
\end{align}

According to the above equation, the redundant feature signals accumulated over the first \(t-1\) time-steps are canceled out, and the static bias term \(b^l\) is also canceled out during the difference operation. At this point, only the current time step’s input \(\hat{x}^l[t] = W^l \hat{x}^{l-1}[t]\), which is passed down from the previous layer, is retained.
\end{proof}

\subsubsection{Proof of \cref{thm: tpc for single-input nonlinear layers}}
\label{supp: proof of the theorem 3.4.}

\begin{customthm}{3.4}
Let \(f\) be a nonlinear layer of order \(l\) that takes only a single input \(x^{l-1}[t]\), such as ReLU, SiLU, MaxPool, LayerNorm, or Softmax. In the TPC framework, if the reconstructed signal \(\hat x^l[t]\) generated by this layer at time \(t\) satisfies the following equation:
\begin{equation}
    \hat{x}^l[t] = t \cdot f(r^{l-1}[t]) - (t-1) \cdot f(r^{l-1}[t-1]),
\end{equation}
then, after any \(t\) time-steps, the steady-state expectation \(r^l[t]\) of the output from this nonlinear layer equals the absolute nonlinear activation value of the source ANN, i.e., \(r^l[t] = f(r^{l-1}[t])\).
\end{customthm}

\begin{proof}
According to Definition \ref{def: temporal predictive coding}, the steady-state expectation \(r^l[t]\) of layer \(l\) at time \(t\) is defined as the time-weighted average of all reconstructed signals \(\hat x^l[t]\):
\begin{equation}
    r^l[t]=\frac{1}{t}\sum^t_{i=1}\hat x^l[i].
\end{equation}
Substitute the equation from the \cref{thm: tpc for single-input nonlinear layers}, \(\hat{x}^l[i] = i \cdot f(r^{l-1}[i]) - (i-1) \cdot f(r^{l-1}[i-1])\) into the above expression and evaluate the sum:
\begin{equation}
    r^l[t] = \frac{1}{t} \sum_{i=1}^t \Big[ i \cdot f(r^{l-1}[i]) - (i-1) \cdot f(r^{l-1}[i-1]) \Big].
\end{equation}
Expand the summand term term by term along the time step \(i\) (assuming the initial steady-state condition \(f(r^{l-1}[0]) = 0\)):
\begin{align}
\sum_{i=1}^t \hat{x}^l[i] &= \Big[ 1 \cdot f(r^{l-1}[1]) - 0 \Big] \notag \\ &+ \Big[ 2 \cdot f(r^{l-1}[2]) - 1 \cdot f(r^{l-1}[1]) \Big] \notag \\ &+ \dots \notag \\ &+ \Big[ t \cdot f(r^{l-1}[t]) - (t-1) \cdot f(r^{l-1}[t-1]) \Big].
\end{align}
With the exception of the last term, the historical nonlinearities at all intermediate time-steps are canceled out. Therefore, the overall result depends only on the current time:
\begin{equation}
    \sum_{i=1}^t \hat{x}^l[i] = t \cdot f(r^{l-1}[t]).
\end{equation}
Substituting this back into the steady-state expectation formula yields:
\begin{equation}
    r^l[t] = \frac{1}{t} \Big[ t \cdot f(r^{l-1}[t]) \Big] = f(r^{l-1}[t]).
\end{equation}
This demonstrates that, by precalculating the target response within the domain of interest and applying temporal differentiation, TPC achieves a lossless transformation for any single-input nonlinear layer.
\end{proof}

\subsubsection{Proof of \cref{thm: tpc for multi-input nonlinear layers}}
\label{supp: proof of the theorem 3.5.}

\begin{customthm}{3.5}
Let \(\otimes\) be an operator with two inputs, \(x^{l-1}_A[t]\) and \(x^{l-1}_B[t]\), such as matrix multiplication or element-wise multiplication. In the ANN-to-SNN conversion, the mapping from the operator \(\otimes\) to the TPC dynamics is given by
\begin{align}
\label{equ: tpc for multi-input nonlinear layer1}
& r_A^{l-1}[t] = \frac{t-1}{t}r_A^{l-1}[t-1] + \frac{\hat{x}_A^{l-1}[t]}{t},  \\
\label{equ: tpc for multi-input nonlinear layer2}
& r_B^{l-1}[t] = \frac{t-1}{t}r_B^{l-1}[t-1] + \frac{\hat{x}_B^{l-1}[t]}{t}, \\ 
\label{equ: tpc for multi-input nonlinear layer3}
\begin{split}
&\hat{x}^l[t] = \frac{1}{t} \big( \hat{x}_A^{l-1}[t] \otimes \hat{x}_B^{l-1}[t] \big) + \frac{t-1}{t} \Big( \hat{x}_A^{l-1}[t] \otimes r_B^{l-1}[t-1] \\
    &\quad + r_A^{l-1}[t-1] \otimes \hat{x}_B^{l-1}[t] - r_A^{l-1}[t-1] \otimes r_B^{l-1}[t-1] \Big),
\end{split}
\end{align}
where \(\hat x^{l-1}_A[t]\) and \(\hat x^{l-1}_B[t]\) are the posterior reconstruction signals of the previous layer received at time step \(t\), respectively; \(r_A^{l-1}[t]\) and \(r_B^{l-1}[t]\) are the steady-state expectations of the internal tracking. The output of layer \(l\) at time-step \(t\) (which serves as the input to layer \(l+1\)) is given by \(x^l[t] = \mathcal{Q}\big(\hat{x}^l[t] - \alpha \cdot \hat{x}^l[t-1]\big)\).
\end{customthm}

\begin{proof}
For the two-input operator \(\otimes\), our goal is to ensure that its time-domain expectation is strictly equivalent to the target mapping of the source ANN, that is:
\begin{equation}
    r^l[t]=r^{l-1}_A[t] \otimes r^{l-1}_B[t].
\end{equation}
According to \cref{thm: tpc for single-input nonlinear layers}, in order to achieve the mapping relationship described above, the ideal reconstructed signal \(\hat{x}^l[t]\) of the current layer must satisfy a strict temporal difference form:
\begin{equation}
    \hat{x}^l[t] = t \cdot \big(r_A^{l-1}[t] \otimes r_B^{l-1}[t]\big) - (t-1) \cdot \big(r_A^{l-1}[t-1] \otimes r_B^{l-1}[t-1]\big).
\end{equation}
According to Definition \ref{def: temporal predictive coding}, we have:
\begin{align}
& t \cdot r_A^{l-1}[t] = (t-1)r_A^{l-1}[t-1] + \hat{x}_A^{l-1}[t], \\
& t \cdot r_B^{l-1}[t] = (t-1)r_B^{l-1}[t-1] + \hat{x}_B^{l-1}[t].
\end{align}
Substituting into \(t \cdot (r_A^{l-1}[t] \otimes r_B^{l-1}[t])\):
\begin{align}
& t\cdot\bigl(
r_A^{l-1}[t]\otimes r_B^{l-1}[t]
\bigr) \notag\\
&\quad= \frac{1}{t}
\bigl(t\cdot r_A^{l-1}[t]\bigr)
\otimes
\bigl(t\cdot r_B^{l-1}[t]\bigr) \notag\\
&\quad= \frac{1}{t}
\bigl(
(t-1)r_A^{l-1}[t-1]+\hat{x}_A^{l-1}[t]
\bigr) \notag\\
&\qquad{}\otimes
\bigl(
(t-1)r_B^{l-1}[t-1]+\hat{x}_B^{l-1}[t]
\bigr).
\end{align}
Expanding fully yields:
\begin{align}
& t\cdot\bigl(
r_A^{l-1}[t]\otimes r_B^{l-1}[t]
\bigr) \notag\\
&\quad= \frac{(t-1)^2}{t}
\bigl(
r_A^{l-1}[t-1]\otimes r_B^{l-1}[t-1]
\bigr) \notag\\
&\qquad{}+ \frac{t-1}{t}
\bigl(
\hat{x}_A^{l-1}[t]\otimes r_B^{l-1}[t-1]
\bigr) \notag\\
&\qquad{}+ \frac{t-1}{t}
\bigl(
r_A^{l-1}[t-1]\otimes\hat{x}_B^{l-1}[t]
\bigr) \notag\\
&\qquad{}+ \frac{1}{t}
\bigl(
\hat{x}_A^{l-1}[t]\otimes\hat{x}_B^{l-1}[t]
\bigr).
\end{align}
The difference equation for the target signal \(\hat{x}^l[t]\), i.e., the expected inertia from the previous time step \((t-1) \cdot \big(r_A^{l-1}[t-1] \otimes r_B^{l-1}[t-1]\big)\):
\begin{align} 
\hat{x}^l[t] &= \frac{(t-1)^2 - t(t-1)}{t} \big( r_A^{l-1}[t-1] \otimes r_B^{l-1}[t-1] \big) \notag \\ 
&+ \frac{t-1}{t} \big( \hat{x}_A^{l-1}[t] \otimes r_B^{l-1}[t-1] \big) \notag \\ 
&+ \frac{t-1}{t} \big( r_A^{l-1}[t-1] \otimes \hat{x}_B^{l-1}[t] \big) \notag \\ 
&+ \frac{1}{t} \big( \hat{x}_A^{l-1}[t] \otimes \hat{x}_B^{l-1}[t] \big). 
\end{align}
Since the cross-term coefficient is \(\frac{(t-1)^2 - t(t-1)}{t} = \frac{1-t}{t} = -\frac{t-1}{t}\), factoring this term yields an analytical solution that completely decouples multivariate matrix multiplication in time and space:
\begin{align}
\hat{x}^l[t]
&= \frac{
\hat{x}_A^{l-1}[t]\otimes\hat{x}_B^{l-1}[t]
}{t} \notag\\
&\quad{}+\frac{t-1}{t}\Bigl(
\hat{x}_A^{l-1}[t]\otimes r_B^{l-1}[t-1]
\notag\\
&\qquad{}+
r_A^{l-1}[t-1]\otimes\hat{x}_B^{l-1}[t]
\notag\\
&\qquad{}-
r_A^{l-1}[t-1]\otimes r_B^{l-1}[t-1]
\Bigr).
\end{align}
It follows that, within the TPC framework, dense operations can be equivalently transformed into sparse tensor multiplication and addition operations. The integration of the network over the time dimension satisfies \(r^l[t] = r_A^{l-1}[t] \otimes r_B^{l-1}[t]\), thus proving this theorem.
\end{proof}

\subsubsection{Proof of \cref{thm: expected quantization noise minimization and damping control}}
\label{supp: proof of the theorem 3.6.}

\begin{theorem}
Let the candidate discrete spike set for a neuron in the \(l\)-th layer be a strictly symmetric exponentially decaying set \(\Lambda^l = \{\lambda^l_1, \dots, \lambda^l_n\} \cup \{-\lambda^l_1, \dots, -\lambda^l_n\} \cup \{0\}\), where adjacent thresholds of the same polarity satisfy \(\lambda^l_{i-1} = 2\lambda^l_i\). Under the assumption of locally uniformly distributed residuals, to minimize the mean squared quantization error \(E = \mathbb{E}\big[(V - \mathcal{Q}(V))^2\big]\) between the input membrane potential \(V\) and the actual emitted spike \(x\), the analytically optimal decision boundary of the quantization function \(\mathcal{Q}(\cdot)\) must be defined as:
\begin{equation}
    \mathcal{Q}(V) = \text{sgn}(V) \cdot \lambda^l_i, \quad \text{if } |V| \in [\beta^* \lambda^l_i, \beta^* \lambda^l_{i-1}),
\end{equation}
where the theoretical optimal decision offset coefficient is \(\beta^* \equiv 0.75\).
\end{theorem}

\begin{proof}
Given the high symmetry of the threshold set, we provide the proof solely for the positive semi-axis \(V > 0\). Let \(B_i\) denote the optimal decision boundary to be determined between adjacent thresholds \(\lambda^l_i\) and \(\lambda^l_{i-1}\). When the input residual \(V \in [\lambda^l_i, \lambda^l_{i-1}]\), the integral of its local mean squared quantization error is defined as: 
\begin{equation}
    E(B_i) = \int_{\lambda^l_i}^{B_i} (v - \lambda^l_i)^2 p(v) dv + \int_{B_i}^{\lambda^l_{i-1}} (v - \lambda^l_{i-1})^2 p(v) dv.
\end{equation}

At low time-steps where prior statistics are lacking, we assume that the prediction error approximately follows a uniform distribution within the local quantization interval, i.e., \(p(v) \approx C\). To find the extremum, we take the first derivative with respect to the boundary \(B_i\) and set it to zero: 
\begin{equation}
    \frac{d E(B_i)}{d B_i} = C \cdot (B_i - \lambda^l_i)^2 - C \cdot (B_i - \lambda^l_{i-1})^2 = 0.
\end{equation}

Solving this equation yields the unique stationary point (minimum): \(B_i^* = \frac{\lambda^l_i + \lambda^l_{i-1}}{2}\), which indicates that the optimal decision boundary must be strictly located at the arithmetic mean of the adjacent thresholds. Substituting the exponentially decaying geometric constraint \(\lambda^l_{i-1} = 2\lambda^l_i\), we obtain:
\begin{equation}
    B_i^* = \frac{\lambda^l_i + 2\lambda^l_i}{2} = 1.5 \lambda^l_i = 0.75 \lambda^l_{i-1}.
\end{equation}

Given that \(|V| \in [\beta \lambda^l_i, \beta \lambda^l_{i-1})\), its upper bound corresponds precisely to the decision boundary. From this, the analytical optimal coefficient \(\beta^* = 0.75\) can be rigorously derived.
\end{proof}

\subsection{Additional Experiment Results}
\label{supp: additional experiment results}

\subsubsection{Component-wise Ablation against DCGS}
\label{supp: component-wise ablation against dcgs}

Table~\ref{tab:tpc_dcgs_ablation} reports a sequential ablation on ImageNet-1K with EVA02-B at \(T=2\) and \(T=4\). Starting from DCGS, introducing the temporal predictor \(P_t\) with \(\alpha=1\) (no damping) improves Top-1 accuracy by \(1.52\) and \(1.17\) percentage points at \(T=2\) and \(T=4\), respectively. Setting \(\alpha=0.7\) adds \(1.50\) and \(1.60\) points. Replacing standard rounding (\(\beta=0.5\)) with the derived decision boundary (\(\beta=0.75\)) yields a further \(0.67\) and \(1.64\) points. Each modification improves accuracy at both time-step budgets in this sequence. The full TPC configuration reaches \(86.29\%\) at \(T=2\) and \(89.83\%\) at \(T=4\), exceeding DCGS by \(3.69\) and \(4.41\) percentage points at the respective time-step budgets.

\begin{table}[htbp]
    \centering
    \vspace{-2mm}
    \caption{Sequential component ablation on ImageNet-1K with EVA02-B. Results are Top-1 accuracy (\%). Each row after DCGS introduces one change to the preceding configuration.}
    \vspace{-2mm}
    \label{tab:tpc_dcgs_ablation}
    \small
    \setlength{\tabcolsep}{4pt}
    \begin{tabular}{@{}lccccc@{}}
        \toprule
        \textbf{Variant}
        & \(P_t\)
        & \(\alpha\)
        & \(\beta\)
        & \(T=2\)
        & \(T=4\) \\
        \midrule
        DCGS
        & \(\times\) & --  & 0.50 & 82.60 & 85.42 \\
        + Prediction
        & \(\checkmark\) & 1.0 & 0.50 & 84.12 & 86.59 \\
        + Damping
        & \(\checkmark\) & 0.7 & 0.50 & 85.62 & 88.19 \\
        + Decision boundary (TPC)
        & \(\checkmark\) & 0.7 & 0.75 & \textbf{86.29} & \textbf{89.83} \\
        \bottomrule
    \end{tabular}
    \vspace{-3mm}
\end{table}

\subsubsection{Conversion of LLaMA-2-7B on WikiText-2}
\label{supp:tpc_llama2_wikitext2}

We evaluate TPC on LLaMA-2-7B to assess its conversion fidelity on a 7B-parameter autoregressive language model. We convert the model without updating its pretrained weights and report perplexity on WikiText-2 at \(T=4\) and \(T=8\). As shown in ~\cref{tab:tpc_llama2_wikitext2}, the source ANN achieves a perplexity of \(5.47\), and the converted model reaches \(6.32\) with four time steps. Increasing the budget to eight time-steps lowers perplexity by \(0.70\) to \(5.62\), narrowing the gap to the source ANN from \(0.85\) to \(0.15\).

\begin{table}[htbp]
    \centering
    \vspace{-2mm}
    \caption{Conversion results for LLaMA-2-7B on WikiText-2. Perplexity (PPL) is lower-is-better. \(\Delta\mathrm{PPL}\) is the increase relative to the source ANN.}
    \vspace{-2mm}
    \label{tab:tpc_llama2_wikitext2}
    \small
    \setlength{\tabcolsep}{5pt}
    \begin{tabular}{@{}lccc@{}}
        \toprule
        \textbf{Configuration}
        & \(\boldsymbol{T}\)
        & \textbf{PPL}\(\downarrow\)
        & \(\Delta\mathrm{PPL}\) \\
        \midrule
        Source ANN       & -- & 5.47 & \(0.00\) \\
        TPC (ANN-to-SNN) & 4  & 6.32 & \(+0.85\) \\
        TPC (ANN-to-SNN) & 8  & 5.62 & \(+0.15\) \\
        \bottomrule
    \end{tabular}
    \vspace{-3mm}
\end{table}

\subsubsection{Comparison with Additional Training-Free Baselines}
\label{supp:tpc_additional_baselines}

We compare TPC with AAR \cite{cao2026towards} on ImageNet-1K and with SpikedAttention \cite{hwang2024spikedattention} on GLUE, as summarized in ~\cref{tab:tpc_additional_baselines}. On ImageNet-1K, TPC achieves \(84.33\%\) and \(85.62\%\) Top-1 accuracy on ViT-B and ViT-L, respectively, using four time-steps. These results exceed those of AAR at \(T=64\) by \(1.50\) and \(0.10\) percentage points, respectively.

On GLUE, TPC at \(T=8\) scores higher on six of the seven tasks than SpikedAttention \cite{hwang2024spikedattention} at \(T=16\) or \(T=24\). The largest gain is on STS-B, where the score increases from \(83.9\) to \(88.8\). On MNLI, TPC scores \(84.2\), compared with \(84.4\) for SpikedAttention. Under the same operation-based energy model, TPC reduces the estimated energy from \(82.1\) to \(57.4\) mJ on QQP and from \(77.5\) to \(55.3\) mJ on MNLI, corresponding to reductions of \(30.1\%\) and \(28.6\%\), respectively.

\begin{table}[t]
    \centering
    \vspace{-2mm}
    \caption{Comparison with AAR and SpikedAttention. \(T\) denotes the number of simulation time-steps. Bold indicates the better result between the two conversion methods in each row.}
    \vspace{-2mm}
    \label{tab:tpc_additional_baselines}
    \small
    \setlength{\tabcolsep}{4pt}
    \begin{tabular*}{\columnwidth}{@{\extracolsep{\fill}}lccccc@{}}
        \toprule
        \multicolumn{6}{@{}l}{\textbf{\textit{(a) ImageNet-1K Top-1 accuracy (\%)}}} \\
        \addlinespace[2pt]
        & & \multicolumn{2}{c}{\textbf{AAR}}
          & \multicolumn{2}{c}{\textbf{TPC}} \\
        \cmidrule(lr){3-4}\cmidrule(lr){5-6}
        \textbf{Backbone} & \textbf{ANN}
        & \(T\) & Acc.
        & \(T\) & Acc. \\
        \midrule
        ViT-B & 84.54 & 64 & 82.83 & 4 & \textbf{84.33} \\
        ViT-L & 85.84 & 64 & 85.52 & 4 & \textbf{85.62} \\
        \midrule
        \multicolumn{6}{@{}l}{\textbf{\textit{(b) GLUE scores with BERT}}} \\
        \addlinespace[2pt]
        & & \multicolumn{2}{c}{\textbf{SpikedAttention}}
          & \multicolumn{2}{c}{\textbf{TPC}} \\
        \cmidrule(lr){3-4}\cmidrule(lr){5-6}
        \multicolumn{2}{@{}l}{\textbf{Task}}
        & \(T\) & Score
        & \(T\) & Score \\
        \midrule
        QQP   & & 24 & 90.8          & 8 & \textbf{90.9} \\
        MNLI  & & 24 & \textbf{84.4} & 8 & 84.2 \\
        SST-2 & & 16 & 92.1          & 8 & \textbf{92.7} \\
        QNLI  & & 24 & 91.0          & 8 & \textbf{91.5} \\
        RTE   & & 16 & 65.0          & 8 & \textbf{65.4} \\
        MRPC  & & 16 & 84.1          & 8 & \textbf{84.9} \\
        STS-B & & 24 & 83.9          & 8 & \textbf{88.8} \\
        \midrule
        \multicolumn{6}{@{}l}{\textbf{\textit{(c) Estimated BERT energy (mJ)}}} \\
        \addlinespace[2pt]
        \multicolumn{2}{@{}l}{\textbf{Task}}
        & \multicolumn{2}{c}{\textbf{SpikedAttention}}
        & \multicolumn{2}{c}{\textbf{TPC}} \\
        \midrule
        \multicolumn{2}{@{}l}{QQP}
        & \multicolumn{2}{c}{82.1}
        & \multicolumn{2}{c}{\textbf{57.4}} \\
        \multicolumn{2}{@{}l}{MNLI}
        & \multicolumn{2}{c}{77.5}
        & \multicolumn{2}{c}{\textbf{55.3}} \\
        \bottomrule
    \end{tabular*}
    \vspace{-3mm}
\end{table}

\subsubsection{Module-Wise Conversion Analysis of Qwen2-VL}
\label{supp:tpc_qwen2vl_module_diagnosis}

We examine which components of Qwen2-VL-7B are most sensitive to TPC conversion at \(T=4\). We convert the vision encoder, cross-modal projector, and language model one at a time, keeping the remaining modules in FP16. We evaluate each configuration on MMStar and MME alongside the full FP16 and fully converted models.

As shown in \cref{tab:tpc_qwen2vl_module_diagnosis}, converting only the vision encoder reduces MMStar and MME by \(0.58\) and \(11.40\) points, respectively. Converting only the projector reduces them by \(0.25\) and \(5.70\) points. The language model is the most sensitive module. Converting it alone reduces MMStar by \(5.50\) points and MME by \(85.30\) points, about \(90\%\) of the score drop observed with full conversion on each benchmark.

\begin{table}[t]
    \centering
    \vspace{-2mm}
    \caption{Module-wise conversion of Qwen2-VL-7B at \(T=4\). Modules outside the listed configuration are kept in FP16. \(\Delta\) denotes the score change relative to the full FP16 model.}
    \vspace{-2mm}
    \label{tab:tpc_qwen2vl_module_diagnosis}
    \small
    \setlength{\tabcolsep}{4pt}
    \begin{tabular}{@{}lrrrr@{}}
        \toprule
        \multicolumn{1}{@{}c}{\multirow{2}{*}{\textbf{Converted modules}}}
        & \multicolumn{2}{c}{\textbf{MMStar}}
        & \multicolumn{2}{c}{\textbf{MME}} \\
        \cmidrule(lr){2-3}\cmidrule(lr){4-5}
        & Score & \(\Delta\)
        & Score & \(\Delta\) \\
        \midrule
        None (FP16 ANN) & 60.70 & --          & 2326.80 & -- \\
        Vision encoder  & 60.12 & \(-0.58\)  & 2315.40 & \(-11.40\) \\
        Projector       & 60.45 & \(-0.25\)  & 2321.10 & \(-5.70\) \\
        Language model  & 55.20 & \(-5.50\)  & 2241.50 & \(-85.30\) \\
        All modules     & 54.60 & \(-6.10\)  & 2232.29 & \(-94.51\) \\
        \bottomrule
    \end{tabular}
    \vspace{-3mm}
\end{table}

\subsection{Limitations}
\label{supp: limitations}

While the proposed Temporal Predictive Coding (TPC) framework significantly advances the training-free conversion from large ANNs to SNNs, it has certain limitations in its current form.

Currently, the TPC framework has been primarily designed and rigorously validated on dense architectures, particularly the standard dense Transformer models (e.g., ViT, EVA02, LlaVA and Qwen-VL). However, modern Large Language Models (LLMs) and Vision-Language Models (VLMs) are increasingly adopting sparse architectures, such as the Mixture-of-Experts (MoE) mechanism, to decouple computational cost from model capacity. The dynamic routing and conditional computation inherent in MoE architectures pose new challenges for SNN conversion, as the sparse activation patterns of MoE routers and the discrete spike firing dynamics of SNNs are conceptually distinct and require novel spatiotemporal mapping strategies.

In our future work, we plan to address this limitation by extending the TPC framework to accommodate MoE-based sparse architectures. Specifically, we aim to explore spike-driven routing mechanisms and asynchronous expert activation strategies. By integrating the macroscopic sparsity of MoE with the microscopic temporal sparsity of SNNs, we hope to unlock even greater energy efficiency and scalability for neuromorphic models in the era of foundation models.

\end{document}